\documentclass[letterpaper]{article}
\usepackage{aaai2026}

\usepackage{times}
\usepackage{helvet}
\usepackage{natbib}
\usepackage{courier}
\usepackage{caption}
\usepackage{graphicx}
\usepackage{newfloat}
\usepackage{listings}
\usepackage{algorithm}
\usepackage{algorithmic}
\usepackage[hyphens]{url}

\usepackage{bm}
\usepackage{tikz}
\usepackage{float}
\usepackage{xparse}
\usepackage{amsmath}
\usepackage{amssymb}
\usepackage{amsfonts}
\usepackage{enumitem}
\usepackage{booktabs}
\usepackage{multirow}
\usepackage{booktabs}
\usepackage{subcaption}
\usepackage[table]{xcolor}

\colorlet{agentblue}{blue!70}
\colorlet{agentbluedark}{blue!60!black}
\colorlet{agentbluelight}{blue!10}
\colorlet{agentblueborder}{blue!65!black}
\colorlet{agentred}{red!65}
\colorlet{agentreddark}{red!55!black}
\colorlet{agentredlight}{red!10}
\colorlet{agentredborder}{red!60!black}

\definecolor{Gray}{gray}{0.92}
\definecolor{lightgray}{HTML}{EFEFEF}
\definecolor{nearwhite}{HTML}{FEFEFE}

\graphicspath{{figures/}}

\usetikzlibrary{arrows.meta, positioning, calc, backgrounds, patterns}

\DeclareCaptionStyle{ruled}{labelfont=normalfont,labelsep=colon,strut=off}

\floatstyle{ruled}
\newfloat{listing}{tb}{lst}{}
\floatname{listing}{Listing}

\title{Large Neighborhood Search for Multi-Agent Task Assignment and Path Finding with Precedence Constraints}
\author{
    Viraj Parimi\textsuperscript{\rm 1},
    Brian Williams\textsuperscript{\rm 1}
}
\affiliations{
    \textsuperscript{\rm 1}
    Massachusetts Institute of Technology \\
    vparimi@mit.edu, williams@csail.mit.edu
}

\begin{document}

\maketitle

\begin{abstract}
Many multi-robot applications require tasks to be completed efficiently and in the correct order, so that downstream operations can proceed at the right time. Multi-agent path finding with precedence constraints (MAPF-PC) is a well-studied framework for computing collision-free plans that satisfy ordering relations when task sequences are fixed in advance. In many applications, however, solution quality depends not only on how agents move, but also on which agent performs which task. This motivates the lifted problem of task assignment and path finding with precedence constraints (TAPF-PC), which extends MAPF-PC by jointly optimizing assignment, precedence satisfaction, and routing cost. To address the resulting coupled TAPF-PC search space, we develop a large neighborhood search approach that starts from a feasible MAPF-PC seed and iteratively improves it through reassignment-based neighborhood repair, restoring feasibility within each selected neighborhood. Experiments across multiple benchmark families and scaling regimes show that the best-performing configuration improves 89.1\% of instances over fixed-assignment seed solutions, demonstrating that large neighborhood search effectively captures the gains from flexible reassignment under precedence constraints.
\end{abstract}

\section{Introduction}
\label{sec:intro}

Multi-agent path finding (MAPF) is a widely studied model for coordinating multiple agents in shared environments. In its classical form, MAPF focuses on computing collision-free paths for agents whose destinations are fixed in advance~\cite{mapf_benchmarks}. This formulation is commonly applied in domains such as warehouse automation and factory logistics, where teams of robots must move efficiently through shared spaces~\cite{warehouse_automation,li2021rhcr}. In many practical settings, however, deciding \emph{which} agent should perform \emph{which} task is itself a key part of the planning problem, and solution quality depends as much on task allocation as on routing.

This coupling becomes even more important when tasks are not independent. In many applications, one task or stage cannot begin until another has been completed. Such precedence relations arise naturally in pickup-and-delivery~\cite{sven-mapd-2017,Liu2019}, warehouse operations~\cite{warehouse_automation,mapf-pc-zhang}, and manufacturing settings such as robotic assembly~\cite{brown-task-path-stanford}. For example, a robot may need to deliver a package before another robot can collect it, or complete multiple sub-assemblies before they can be combined~\cite{sven-mapd-2017,brown-task-path-stanford}. Once such dependencies are present, task assignment, temporal ordering, and collision-free routing become tightly coupled, since changing the assignment affects who goes where, when downstream tasks become available, and how agents interact in shared space~\cite{mapf-pc-zhang,brown-task-path-stanford}.

Prior work has explored several ways of integrating assignment and routing in multi-agent systems, sometimes including local task ordering, but most do not treat precedence relations as explicit global constraints. In one line of work, agents are jointly assigned to targets while collision-free paths are planned simultaneously, as in TAPF~\cite{ma2016tapf} and Conflict-Based Search with Task Assignment (CBS-TA)~\cite{hoenig2018cbsta}. Other formulations enrich the task model by introducing additional temporal or structural requirements. Multi-Goal TAPF (MG-TAPF) extends this to \emph{multi-goal tasks} with prescribed within-task goal sequences~\cite{zhong2022mgtapf}, while Multi-Agent Pickup-and-Delivery (MAPD) addresses a lifelong regime in which pickup-and-delivery tasks arrive online and must be assigned and routed over time~\cite{sven-mapd-2017,Liu2019}. Together, these formulations show that assignment and motion often need to be optimized jointly, but none of them combines explicit cross-task precedence constraints with flexible assignment.

This gap becomes sharper when precedence relations are themselves explicit planning constraints. MAPF-PC is an established framework for multi-agent planning with precedence constraints in the fixed-assignment setting, where each agent is given a goal sequence and the planner must compute collision-free paths that satisfy precedence relations among the goals~\cite{mapf-pc-zhang}. Related formulations that combine flexible assignment with precedence constraints have been explored in robotic assembly and manufacturing~\cite{brown-task-path-stanford,liu2024pctapf}, but these approaches rely on exact search or sequential greedy insertion and have not been demonstrated at the scale of hundreds of agents and thousands of tasks. In this paper, we study task assignment and path finding with precedence constraints (TAPF-PC), treating it as a lifted counterpart of MAPF-PC. While MAPF-PC addresses routing and precedence satisfaction for a given assignment, TAPF-PC must additionally reason over assignment choices. Figure~\ref{fig:fixed-vs-flexible-counterexample} illustrates this distinction with a simple counterexample, showing that fixed-assignment optimality does not resolve TAPF-PC once reassignment is allowed. Accordingly, we view TAPF-PC as a search problem above MAPF-PC, in which the outer search explores reassignments while MAPF-PC methods repair the induced fixed-assignment neighborhoods.

The resulting search space couples assignment, precedence satisfaction, and collision avoidance, making direct optimization challenging. Exact joint methods are available for several related assignment-and-routing formulations~\cite{ma2016tapf,hoenig2018cbsta,zhong2022mgtapf,brown-task-path-stanford}, but the scale of the lifted TAPF-PC search space motivates scalable suboptimal search. To this end, we develop a large neighborhood search (LNS) framework for TAPF-PC that starts from a feasible MAPF-PC seed and explores alternative assignments around it. At each iteration, the method selects a precedence-aware task neighborhood for reconsideration, keeps the exterior solution fixed, and repairs the affected region under the timing and routing constraints induced across the neighborhood boundary. In this way, fixed-assignment MAPF-PC methods~\cite{mapf-pc-zhang} are embedded as repair engines within a higher-level search over assignment and routing.

Our contributions are twofold. First, we develop a large neighborhood search framework for TAPF-PC that decomposes the problem into an outer reassignment search with precedence-aware neighborhood destruction and inner MAPF-PC repair subproblems. Second, through systematic experiments across multiple benchmark families and scaling regimes, we show that flexible reassignment improves 89.1\% of instances and characterize the repair and neighborhood design choices that drive these results.

\begin{figure}[t]
\centering
\resizebox{0.9\columnwidth}{!}{%
\begin{tikzpicture}[
    x=0.425cm, y=0.425cm,
    >=Latex,
    font=\sffamily\tiny,
    start/.style={
        draw=none, fill=#1, rectangle,
        minimum size=3.2mm, inner sep=0pt,
        rounded corners=0.6mm
    },
    start/.default={black},
    goalblue/.style={
        draw=agentblueborder, fill=agentbluelight, circle,
        line width=0.45pt, minimum size=3.6mm, inner sep=0pt,
        font=\sffamily\tiny\bfseries
    },
    goalred/.style={
        draw=agentredborder, fill=agentredlight, circle,
        line width=0.45pt, minimum size=3.6mm, inner sep=0pt,
        font=\sffamily\tiny\bfseries
    },
    goalblack/.style={
        draw=black!65, fill=black!10, circle,
        line width=0.45pt, minimum size=3.6mm, inner sep=0pt,
        font=\sffamily\tiny\bfseries
    },
    blocked/.style={fill=black!75, draw=black!85, line width=0.3pt},
    blockedx/.style={line width=0.35pt, black!40},
    prec/.style={line width=0.55pt, black!45, -{Latex[length=1.1mm, width=0.8mm]}},
    aone/.style={
        line width=1.0pt, blue!60!black,
        -{Latex[length=1.3mm, width=1.0mm]},
        line cap=round, line join=round
    },
    aoneglow/.style={
        line width=2.8pt, blue!25, line cap=round, line join=round
    },
    atwo/.style={
        line width=1.0pt, red!55!black,
        -{Latex[length=1.3mm, width=1.0mm]},
        line cap=round, line join=round
    },
    atwoglow/.style={
        line width=2.8pt, red!20, line cap=round, line join=round
    },
    paneltitle/.style={font=\footnotesize\bfseries, text=black!80},
    costbox/.style={
        draw=black!20, rounded corners=1.2mm, fill=black!2.5,
        inner xsep=3pt, inner ysep=2.2pt, align=center,
        font=\sffamily\tiny
    },
    seclabel/.style={
        font=\sffamily\tiny\bfseries, text=black!50,
        fill=white, inner sep=1pt
    },
    waypoint/.style={circle, fill=#1!50!black, inner sep=0pt, minimum size=1.2pt},
    waypoint/.default={black}
]
 
\begin{scope}[shift={(0, 15.8)}]
    \draw[rounded corners=1.5mm, fill=black!3, draw=black!18, line width=0.35pt]
        (-0.5, -0.75) rectangle (7.5, 0.75);
 
    \node[seclabel] at (3.5, 0.75) {Precedence constraints};
 
    \node[goalblue] (pc1) at (1.2, 0) {$g_1$};
    \node[goalblue] (pc2) at (2.7, 0) {$g_2$};
    \draw[prec] (pc1) -- (pc2);
 
    \node[goalred]  (pc3) at (4.3, 0) {$g_3$};
    \node[goalred]  (pc4) at (5.8, 0) {$g_4$};
    \draw[prec] (pc3) -- (pc4);
\end{scope}
 
\begin{scope}[shift={(0, 9.0)}]
    \node[paneltitle, anchor=south] at (3.5, 4.25) {(a)\; Fixed Assignment (MAPF-PC)};
 
    \begin{scope}[on background layer]
        \fill[white] (0,0) rectangle (7,4);
        \draw[step=1, black!10, line width=0.25pt] (0,0) grid (7,4);
    \end{scope}
    \draw[black!35, line width=0.55pt, rounded corners=0.3mm] (0,0) rectangle (7,4);
 
    \fill[blocked] (2,1) rectangle (3,2);
    \fill[blocked] (3,1) rectangle (4,2);
    \draw[white, line width=0.35pt] (2,1) -- (3,2);  \draw[white, line width=0.35pt] (2,2) -- (3,1);
    \draw[white, line width=0.35pt] (3,1) -- (4,2);  \draw[white, line width=0.35pt] (3,2) -- (4,1);
 
    \begin{scope}[on background layer]
        \draw[aoneglow] (0.5,0.5) -- (0.5,3.5) -- (1.5,3.5) -- (5.5,3.5);
        \draw[aone]     (0.5,0.5) -- (0.5,3.5) -- (1.5,3.5) -- (5.5,3.5);
        \draw[atwoglow] (6.5,0.5) -- (6.5,2.5) -- (5.5,2.5) -- (1.5,2.5);
        \draw[atwo]     (6.5,0.5) -- (6.5,2.5) -- (5.5,2.5) -- (1.5,2.5);
    \end{scope}
 
    \node[waypoint=blue] at (0.5,3.5) {};
    \node[waypoint=blue] at (1.5,3.5) {};
    \node[waypoint=red]  at (6.5,2.5) {};
    \node[waypoint=red]  at (5.5,2.5) {};
 
    \node[start=agentblue]  (s1a) at (0.5,0.5) {};
    \node[left=1.2mm of s1a, font=\sffamily\tiny\bfseries, text=agentbluedark] {$a_1$};
    \node[start=agentred]   (s2a) at (6.5,0.5) {};
    \node[right=1.2mm of s2a, font=\sffamily\tiny\bfseries, text=agentreddark] {$a_2$};
 
    \node[goalblue] at (1.5,3.5) {$g_1$};
    \node[goalblue] at (1.5,2.5) {$g_2$};
    \node[goalred]  at (5.5,2.5) {$g_3$};
    \node[goalred]  at (5.5,3.5) {$g_4$};
 
    \node[costbox] at (3.5, -0.85) {%
        {\color{agentbluedark}$a_1\!:\!(g_1,g_4)$}%
        \;\;%
        {\color{agentreddark}$a_2\!:\!(g_3,g_2)$}%
        \qquad $\boldsymbol{J_0\!=\!15}$%
    };
\end{scope}
 
\begin{scope}[shift={(0, 1.5)}]
    \node[paneltitle, anchor=south] at (3.5, 4.25) {(b)\; Flexible Assignment (TAPF-PC)};
 
    \begin{scope}[on background layer]
        \fill[white] (0,0) rectangle (7,4);
        \draw[step=1, black!10, line width=0.25pt] (0,0) grid (7,4);
    \end{scope}
    \draw[black!35, line width=0.55pt, rounded corners=0.3mm] (0,0) rectangle (7,4);
 
    \fill[blocked] (2,1) rectangle (3,2);
    \fill[blocked] (3,1) rectangle (4,2);
    \draw[white, line width=0.35pt] (2,1) -- (3,2);  \draw[white, line width=0.35pt] (2,2) -- (3,1);
    \draw[white, line width=0.35pt] (3,1) -- (4,2);  \draw[white, line width=0.35pt] (3,2) -- (4,1);
 
    \begin{scope}[on background layer]
        \draw[aoneglow] (0.5,0.5) -- (0.5,3.5) -- (1.5,3.5) -- (1.5,2.5);
        \draw[aone]     (0.5,0.5) -- (0.5,3.5) -- (1.5,3.5) -- (1.5,2.5);
        \draw[atwoglow] (6.5,0.5) -- (6.5,2.5) -- (5.5,2.5) -- (5.5,3.5);
        \draw[atwo]     (6.5,0.5) -- (6.5,2.5) -- (5.5,2.5) -- (5.5,3.5);
    \end{scope}
 
    \node[waypoint=blue] at (0.5,3.5) {};
    \node[waypoint=blue] at (1.5,3.5) {};
    \node[waypoint=red]  at (6.5,2.5) {};
    \node[waypoint=red]  at (5.5,2.5) {};
 
    \node[start=agentblue]  (s1b) at (0.5,0.5) {};
    \node[left=1.2mm of s1b, font=\sffamily\tiny\bfseries, text=agentbluedark] {$a_1$};
    \node[start=agentred]   (s2b) at (6.5,0.5) {};
    \node[right=1.2mm of s2b, font=\sffamily\tiny\bfseries, text=agentreddark] {$a_2$};
 
    \node[goalblue] at (1.5,3.5) {$g_1$};
    \node[goalblue] at (1.5,2.5) {$g_2$};
    \node[goalred]  at (5.5,2.5) {$g_3$};
    \node[goalred]  at (5.5,3.5) {$g_4$};
 
    \node[costbox] at (3.5, -0.85) {%
        {\color{agentbluedark}$a_1\!:\!(g_1,g_2)$}%
        \;\;%
        {\color{agentreddark}$a_2\!:\!(g_3,g_4)$}%
        \qquad $\boldsymbol{J_1\!=\!9}$%
    };
\end{scope}
 
\end{tikzpicture}%
}
 
 
{%
\centering
\scriptsize
\begin{tikzpicture}[baseline=-0.5ex]
  \fill[agentblue] (0,0) rectangle (1.5mm,1.5mm);
\end{tikzpicture}\;Start
\qquad
\begin{tikzpicture}[baseline=-0.5ex]
  \draw[black!65, line width=0.35pt] (0.75mm,0.75mm) circle (0.85mm);
  \fill[black!10] (0.75mm,0.75mm) circle (0.85mm);
  \node[font=\tiny\bfseries, text=black!65] at (0.75mm,0.75mm) {$g$};
\end{tikzpicture}\;Goal
\qquad
\begin{tikzpicture}[baseline=-0.5ex]
  \draw[line width=0.4pt, black!45, -{Latex[length=1mm]}] (0,0.75mm) -- (3mm,0.75mm);
\end{tikzpicture}\;Prec.
\qquad
\begin{tikzpicture}[baseline=-0.5ex]
  \fill[black!75] (0,0) rectangle (1.5mm,1.5mm);
  \draw[white, line width=0.2pt] (0,0) -- (1.5mm,1.5mm);
  \draw[white, line width=0.2pt] (0,1.5mm) -- (1.5mm,0);
\end{tikzpicture}\;Wall
\par
}
 
\caption{%
Fixed vs.\ flexible goal assignment under precedence constraints
$(g_1 \!\prec\! g_2,\; g_3 \!\prec\! g_4)$. $J$ denotes sum of path costs; assigned goal sequences are shown below each panel.
\textbf{(a)}~Fixed Assignment (MAPF-PC): $J_0\!=\!15$.
\textbf{(b)}~Flexible Assignment (TAPF-PC): $J_1\!=\!9$, showing that fixed-assignment optimality does not imply TAPF-PC optimality.%
}
\label{fig:fixed-vs-flexible-counterexample}
\end{figure}

\section{Problem Formulation}
\label{sec:problem-formulation}

We model the environment as an undirected graph $G=(V,E)$, where $V$ is the set of vertices and $E$ is the set of edges. Let $A=\{a_1,\dots,a_k\}$ denote a set of $k$ agents, with each agent $a_i$ starting at a distinct vertex $s_i \in V$. Time is discrete, and at each timestep an agent may move to an adjacent vertex or wait at its current vertex. As in standard MAPF, vertex conflicts and edge conflicts are forbidden, meaning that two agents may not occupy the same vertex at the same timestep or traverse the same edge in opposite directions simultaneously~\cite{mapf_benchmarks}.

Let $\Theta=\{\theta_1,\dots,\theta_m\}$ denote the set of $m$ tasks, where each task $\theta \in \Theta$ is associated with a goal vertex $g(\theta)\in V$. A precedence constraint is an ordered pair $(\theta_u,\theta_v)$ indicating that task $\theta_v$ may be completed only after task $\theta_u$ has been completed. Let $P \subseteq \Theta \times \Theta$ denote the set of precedence constraints, and assume that $P$ is acyclic. A task $\theta$ is completed when an agent reaches $g(\theta)$ and all predecessor constraints are satisfied. We now formally state the two problem formulations relevant to this work, namely MAPF-PC, in which the task assignment is fixed, and TAPF-PC, in which it is part of the solution.

\subsection{MAPF-PC}

In MAPF-PC, each agent $a_i$ is given a fixed ordered task sequence $c_i = \langle \theta_{i,1},\theta_{i,2},\dots,\theta_{i,m_i}\rangle$ of $m_i = |c_i|$ tasks as part of the input. Let $C = \{c_1,\dots,c_k\}$ denote the collection of all agent task sequences. Some sequences may be empty, in which case the corresponding agent is assigned no tasks. These sequences form a partition of $\Theta$, so that each task is assigned to exactly one agent in advance and $\sum_{i=1}^{k} m_i = m$.

For a task $\theta$, let $\tau(\theta)$ denote its completion time. A feasible MAPF-PC solution consists of a collision-free path for each agent such that all tasks in its assigned sequence are completed. In addition, the prescribed local order must be respected, so $\tau(\theta_{i,j}) < \tau(\theta_{i,j+1})$ for all consecutive tasks in $c_i$, and every global precedence constraint $(\theta_u,\theta_v)\in P$ must satisfy $\tau(\theta_u) < \tau(\theta_v)$. MAPF-PC therefore extends classical MAPF by allowing each agent to execute a prescribed sequence of tasks under additional global precedence constraints~\cite{mapf_benchmarks,mapf-pc-zhang}.

\subsection{TAPF-PC}

TAPF-PC lifts MAPF-PC by making the task sequences $C=\{c_1,\dots,c_k\}$ part of the solution rather than part of the input~\cite{brown-task-path-stanford,liu2024pctapf}. A feasible TAPF-PC solution must therefore determine a partition of $\Theta$ into per-agent task sequences, the order of tasks within each sequence, and collision-free paths that realize those choices while satisfying all precedence constraints.

Equivalently, a TAPF-PC solution induces a collection of ordered per-agent task sequences $C = \{c_1,\dots,c_k\}$, but unlike MAPF-PC, these sequences are not specified in advance. In this paper, we study TAPF-PC in the offline setting, where all tasks and precedence constraints are known at the beginning of planning. Unless stated otherwise, we seek a feasible TAPF-PC solution of minimum sum of costs, where the cost of an agent is the timestep at which it completes its final assigned task. This objective is standard in MAPF and in joint assignment-and-routing formulations such as CBS-TA~\cite{hoenig2018cbsta}, although TAPF has also been studied under makespan objectives~\cite{ma2016tapf}.

\section{Overall Framework}
\label{sec:overall}

We solve TAPF-PC using a large neighborhood search (LNS) framework that operates on complete feasible solutions. At iteration $t$, the incumbent feasible solution consists of ordered per-agent task sequences $C_t=\{c_1^t,\dots,c_k^t\}$ together with collision-free paths that realize those sequences. The search is initialized from a feasible fixed-assignment MAPF-PC solution (Section~\ref{sec:seed-solution}), obtained using an existing MAPF-PC solver~\cite{mapf-pc-zhang}. Starting from this seed, the framework repeatedly destroys (Section~\ref{sec:neighborhood-destruction}) and repairs precedence-aware neighborhoods (Section~\ref{sec:repair-mechanism}) to explore reassignment opportunities.

Unlike LNS for standard MAPF, our neighborhoods are defined over tasks rather than over agents or paths~\cite{huang2022-mapf-lns,li2022mapf-lns2}. This distinction is essential in TAPF-PC because removing one task may force dependent successor tasks to be reconsidered, while repairing a task may require restoring missing predecessor tasks in the precedence graph to preserve feasibility. Accordingly, each iteration selects a task neighborhood, derives boundary conditions from the frozen exterior, constructs a reassignment-and-order proposal, solves the induced MAPF-PC repair instance, and stitches the result into the incumbent.

This organization yields a two-level architecture. At the outer level, LNS searches over reassignment and local ordering decisions for TAPF-PC. At the inner level, the induced repair instance is solved as a constrained MAPF-PC subproblem embedded in the current global solution. Algorithm~\ref{alg:tapfpc-lns} summarizes the overall control flow of the proposed framework, from seed generation through neighborhood destruction, repair, acceptance, and post-refinement. Here $S$ denotes the current search state, $S^\star$ denotes the best feasible solution found so far, and $\widetilde{S}$ denotes the candidate produced by the current repair step.

\begin{algorithm}[t]
\caption{Large Neighborhood Search for TAPF-PC.}
\label{alg:tapfpc-lns}
\begin{algorithmic}[1]
\REQUIRE Graph $G$, agents $A$, tasks $\Theta$, Precedence $P$
\ENSURE Best feasible TAPF-PC solution found
\STATE Construct a fixed-assignment MAPF-PC seed $S^\star$
\STATE $S \leftarrow S^\star$
\WHILE{search budget remains}
    \STATE Select and expand destroy seed into $\Theta_t^{-}$
    \STATE Construct a reassignment-and-order proposal for $\Theta_t^{-}$
    \STATE Solve the induced local MAPF-PC repair subproblem
    \STATE Stitch repaired paths into exterior as candidate $\widetilde{S}$
    \STATE Validate $\widetilde{S}$ and update $S$ via the acceptance rule
    \IF{$\widetilde{S}$ improves the best feasible solution}
        \STATE $S^\star \leftarrow \widetilde{S}$
    \ENDIF
\ENDWHILE
\STATE Apply post-refinement to $S^\star$
\STATE \textbf{return} $S^\star$
\end{algorithmic}
\end{algorithm}

\section{Large Neighborhood Search for TAPF-PC}
\label{sec:lns-tapf-pc}

\subsection{Seed Solution}
\label{sec:seed-solution}

The search begins from a complete fixed-assignment seed that specifies, for each task, an assigned agent, a within-agent execution order, and explicit paths that realize those choices. To obtain this seed, we first construct a fixed-assignment MAPF-PC instance using a precedence-aware greedy assignment-and-order heuristic. The initializer maintains the set of precedence-ready tasks, whose predecessor tasks in $P$ have already been assigned, and repeatedly assigns one such task to the agent whose current partial sequence has the smallest estimated completion time. Once the per-agent task sequences are formed, the resulting assignment and within-agent order define a fully specified fixed-assignment MAPF-PC instance, which is then solved with a MAPF-PC method, such as Priority-Based Search (PBS-PC) or Conflict-Based Search (CBS-PC), to obtain collision-free paths for all agents~\cite{mapf-pc-zhang}.

\subsection{Neighborhood Destruction}
\label{sec:neighborhood-destruction}

Each LNS iteration begins by selecting a task neighborhood to destroy. The destroy phase selects a seed set of tasks $\widehat{\Theta}_t^{-}$ and transforms it into the final destroyed neighborhood $\Theta_t^{-}$ as described below. The neighborhood-size parameter controls the seed size, but the actual destroyed set may be larger after precedence closure.

\subsubsection{Destroy Operator Families}

The destroy operators used in our framework fall into three broad families. The first family consists of task-seeded operators that adapt standard destroy strategies from prior LNS work, including random removal, cost-based removal, conflict-driven removal, and Shaw-style relatedness removal~\cite{shaw1997lns,alnsused}. 

The second family consists of the precedence-aware operators introduced in this work. The \textsc{Precedence-Wait} operator targets tasks whose execution is delayed by unmet precedence requirements. For an assigned task $\theta$, let $w_{\mathrm{prec}}(\theta)$ denote the realized precedence wait, the gap between the agent's arrival at $g(\theta)$ and the earliest time $\theta$ can begin given its predecessors' completion times. The operator ranks tasks by decreasing $w_{\mathrm{prec}}(\theta)$ and seeds destruction from those with the largest wait, expanding to their immediate precedence neighbors.

While \textsc{Precedence-Wait} targets individual tasks, the \textsc{Low-Slack} operator targets precedence edges directly. For a precedence edge $(\theta_u,\theta_v)\in P$, let $\sigma(\theta_u,\theta_v)$ denote the temporal slack between the completion of $\theta_u$ and the realized start of $\theta_v$ in the current incumbent solution. Small slack indicates a tight precedence handoff, leaving little room to absorb changes in completion time without affecting downstream execution. The \textsc{Low-Slack} operator ranks edges by increasing slack and uses the tasks at both ends of the tightest edges as the destroy seed. Together, these operators steer the search toward precedence bottlenecks that are not visible to cost- or collision-based destroy rules.

The third family is agent-seeded, motivated by localized repair ideas from MAPF-LNS2~\cite{li2022mapf-lns2}. The \textsc{Agent-Conflict} operator seeds destruction from tasks owned by agents involved in recent conflicts, while the \textsc{Failure-Recovery} operator targets tasks owned by agents whose most recent repair attempt failed. Although seeded from agents, the resulting neighborhood is still task-based. Across all three families, the destroy policy may be fixed or selected adaptively using an adaptive large neighborhood search (ALNS) portfolio~\cite{alns}.

\subsubsection{Precedence Closure and Neighborhood Preparation}

Starting from the seed $\widehat{\Theta}_t^{-}$, the algorithm takes the transitive successor closure in the precedence graph, adding every descendant of the seed tasks to the removed set to form the destroyed neighborhood $\Theta_t^{-}$. This closure is essential because once a task is removed, every downstream task that depends on it can no longer be treated as fixed. Predecessor tasks, by contrast, may remain in the exterior solution and later induce boundary conditions for repair.

After successor closure, tasks in $\Theta_t^{-}$ are excised from the current assignment and their path segments are removed. If a removed task is followed by a surviving task on the same agent, that immediate successor becomes a boundary task. Such a task remains assigned, but its incoming path segment from a now-destroyed predecessor has been invalidated and must therefore be recomputed before full repair. Boundary tasks are replanned to account for the missing predecessor, and later surviving tasks are retimed to reflect updated predecessor completion times. In our framework, these updates are computed with Multi-Label A* (MLA*)~\cite{mla_star}. Outside this affected boundary, assignment and path structure remain fixed, and the destroyed set $\Theta_t^{-}$ with the patched exterior context $C_t^{\mathrm{fix}}$, comprising the frozen assignment, paths, and incumbent completion times of all tasks outside the destroyed neighborhood, is passed to the repair stage.

\subsection{Repair Mechanisms}
\label{sec:repair-mechanism}

Given the destroyed task neighborhood $\Theta_t^{-}$ and the fixed exterior context $C_t^{\mathrm{fix}}$, the repair mechanisms described below operate on the per-agent task sequences after removing tasks in $\Theta_t^{-}$. Let $c_i^t$ denote the resulting sequence of agent $a_i$.

\subsubsection{Regret-Based Repair}
\label{sec:regret-repair}

Regret repair rebuilds the destroyed portion of the solution by greedily reinserting destroyed tasks one at a time into the current per-agent task sequences. For a destroyed task $\theta \in \Theta_t^{-}$, a feasible insertion spot is defined by a candidate owner $a_i$ together with an insertion position in $c_i^t$ such that the resulting local order remains consistent with the precedence constraints and the frozen exterior context. Among all feasible insertion spots for $\theta$, let $\Delta_1(\theta) \leq \Delta_2(\theta) \leq \dots$ denote the corresponding insertion costs in increasing order, so that $\Delta_1(\theta)$ is the cost of the best feasible insertion of $\theta$ into the current partial repair. The regret score of task $\theta$ is defined as $\rho(\theta) = \Delta_2(\theta) - \Delta_1(\theta)$, the difference between the best and second-best feasible insertion costs~\cite{alnsused,mapd}. A large regret indicates that failing to place $\theta$ at its best current spot would incur a substantial loss relative to the remaining options.

Repair proceeds greedily, guided by these regret scores. At each step, the algorithm selects the destroyed task with highest regret, inserts it at its best feasible spot, updates the partial solution, and then recomputes the relevant insertion costs for the remaining destroyed tasks. This compute-intensive process continues until all tasks in $\Theta_t^{-}$ have been reinserted or no feasible insertion remains, in which case the repair attempt is abandoned. Notably, regret-based repair determines reassignment and local ordering entirely through greedy insertion scores, so path feasibility enters the repair only through these local insertion evaluations.

\subsubsection{MAPF-PC Neighborhood Repair}
\label{sec:neighborhood-repair}

Let $A_t^{\mathrm{mut}} \subseteq A$ denote the set of agents whose task sequences may change during repair. Before proposal construction begins, the framework computes an initial release bound for each destroyed task $\theta \in \Theta_t^{-}$ from cross-boundary precedence edges. Specifically, fixed predecessors of $\theta$ induce an earliest feasible completion bound
\[
\underline{\tau}(\theta)=1+\max_{(\varphi,\theta)\in P,\ \varphi \notin \Theta_t^{-}}\tau^{\mathrm{inc}}(\varphi),
\]
where $\tau^{\mathrm{inc}}$ denotes the incumbent completion time from the frozen exterior context $C_t^{\mathrm{fix}}$. As each destroyed task is inserted during proposal construction, its approximate completion time becomes available, and the release bounds of any remaining destroyed successors are tightened.

Repair then constructs a local assignment-and-order proposal for the tasks in $\Theta_t^{-}$. The destroyed tasks are processed in a precedence-respecting topological order over the destroyed subset. Among currently ready destroyed tasks, the proposal builder prioritizes tasks with smaller release bounds. This ordering is computed once before the insertion loop and remains fixed throughout proposal construction. For the current task $\theta \in \Theta_t^{-}$, the algorithm considers candidate owners $a_i \in A_t^{\mathrm{mut}}$ together with feasible insertion positions in $c_i^t$. In \emph{local} mode, the candidate owner set is restricted to a bounded mutable-agent neighborhood around the current owners of the destroyed tasks, whereas in \emph{global} mode any agent may be considered as in regret repair.

A candidate insertion of $\theta$ at position $p$ in agent $a_i$'s sequence is admissible if it respects the local same-agent ordering, satisfies the release bound $\underline{\tau}(\theta)$, remains reachable under the current approximate completion-time estimate, and does not conflict with the frozen exterior occupancy reservations. Among admissible positions, the proposal selects the one minimizing a static insertion-detour score based on shortest-path distances $d(\cdot,\cdot)$, ignoring dynamic agent interactions. Exact collision avoidance is deferred to the MAPF-PC repair solve that follows later. Each insertion is committed immediately, after which approximate completion times and release bounds are updated. Repeating this over all tasks in $\Theta_t^{-}$ in topological order yields the proposed mutable task sequences $\widehat{C}_t^{\mathrm{mut}}$. Once $\widehat{C}_t^{\mathrm{mut}}$ has been fixed, we solve a local MAPF-PC subproblem whose task sequences are given by $\widehat{C}_t^{\mathrm{mut}}$ and whose exterior context is inherited from $C_t^{\mathrm{fix}}$~\cite{mapf-pc-zhang}.

\subsection{SIPPS Integration}
\label{sec:sipps-integration}

The effectiveness of neighborhood repair depends not only on the quality of the local reassignment proposal, but also on the speed with which that proposal can be realized as collision-free paths. Because the outer LNS loop invokes repair repeatedly, a slow low-level planner directly reduces the number of neighborhoods that can be explored within a fixed runtime budget. For this reason, we instantiate neighborhood MAPF-PC repair with Safe Interval Path Planning with Soft constraints (SIPPS)~\cite{li2022mapf-lns2} as the low-level path planner rather than the default MLA* planner used by the MAPF-PC solver. This preserves the MAPF-PC repair structure while reducing per-iteration cost. Recall from Section~\ref{sec:neighborhood-repair} that the low-level planner must compute paths for each mutable agent that visit its repaired goal sequence, respect the frozen exterior occupancy reservations, and satisfy the precedence-derived release bounds.

We adapt SIPPS to the precedence-constrained neighborhood-repair setting, following the same stage-indexed approach by which MAPF-PC adapts its low-level planner to ordered goal sequences~\cite{mapf-pc-zhang}. After the proposal fixes the local assignment and order, the resulting temporal constraints are exported to the MAPF-PC repair solver, which derives stage-level lower and upper timing bounds for SIPPS. SIPPS enforces these bounds directly during low-level search. Arrivals before the lower bound are deferred by waiting when possible, whereas arrivals after the upper bound are infeasible. In addition, SIPPS prunes branches whose optimistic completion estimates already violate downstream timing bounds, avoiding search effort on provably infeasible paths.

A further consequence of using SIPPS is that the repaired neighborhood need not be realized only against hard obstacles. SIPPS also supports soft obstacles, in which collisions with the frozen exterior are penalized rather than strictly prohibited. This allows the low-level planner to return a neighborhood solution that remains in limited soft conflict with the frozen exterior solution. These soft conflicts arise only at the boundary between the repaired neighborhood and the fixed world outside it. They do not alter the assignment-and-order proposal itself. This distinction gives rise to two repair modes. In the \emph{hard} mode, only fully conflict-free neighborhood realizations are retained. In the \emph{relaxed} mode, soft-conflict realizations may also be returned, leaving the outer search to resolve the remaining interface conflicts later.

\subsection{Acceptance, Update, and Post-Refinement}
\label{sec:acceptance-update-postrefine}

After neighborhood repair produces a candidate solution, the repaired paths are stitched back into the fixed exterior context and validated against the global precedence constraints and path-conflict conditions before any update is made. The framework supports multiple acceptance rules for updating the search state. In the main configuration studied in this paper, only fully conflict-free repair solutions are retained, and valid candidates are accepted or rejected using threshold acceptance~\cite{thresholdacc}. A relaxed variant permits solutions with soft conflicts at the neighborhood boundary, accepting candidates only if their conflict counts does not worsen. 

After the LNS loop terminates, a post-refinement phase is applied to the best solution found. During the main LNS loop, all MAPF-PC solves operate on local neighborhoods, so the final paths are composed of independently repaired subproblems rather than a globally planned routing. Post-refinement addresses this by treating the assignment and within-agent order discovered by LNS as fixed and invoking a full MAPF-PC solve over the complete instance. The result is accepted only when it yields a valid solution that strictly improves the current best incumbent.

\section{Experimental Setup}
\label{sec:experimental-setup}

Our experiments address the following questions:
\begin{itemize}
\item[\textbf{Q1.}] How do different repair strategies and reassignment scopes compare for TAPF-PC?
\item[\textbf{Q2.}] How does search behavior evolve over time?
\item[\textbf{Q3.}] How does performance vary with team size and precedence density?
\item[\textbf{Q4.}] Are the SoC gains driven primarily by reductions in precedence-induced waiting?
\end{itemize}

\begin{figure*}[t]
    \centering
    \includegraphics[width=\textwidth]{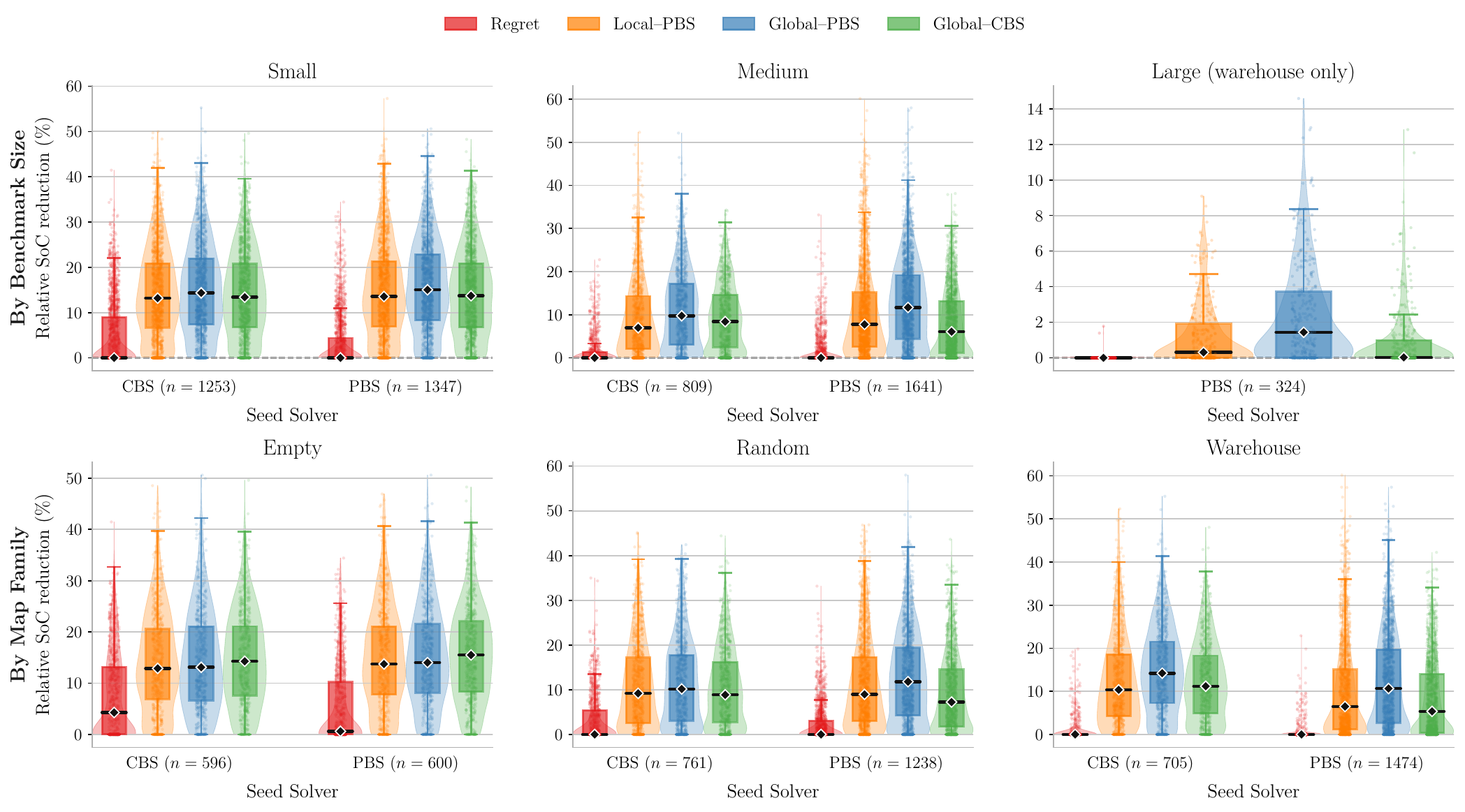}
    \caption{
        Main method comparison across seeded benchmark subsets. Each panel reports relative sum of costs reduction over the fixed-assignment seed (higher is better). \textbf{Top row}: By benchmark tier. \textbf{Bottom row}: By map family.
    }
    \label{fig:main-method-comparison}
\end{figure*}

\subsection{Benchmarks and Instance Generation}
\label{sec:benchmarks}

We evaluate the proposed TAPF-PC methods on a benchmark suite built from four map families, namely \texttt{empty-16-16}, \texttt{empty-32-32}, \texttt{random-32-32-20}, and \texttt{warehouse-10-20-10-2-1}. These correspond to grid maps of size $16 \times 16$, $32 \times 32$, $32 \times 32$, and $161 \times 63$, respectively. Instances are parameterized by the number of agents $k$, tasks $m$, and precedence constraints $|P|$.

The benchmark suite is organized into three tiers, denoted \emph{small}, \emph{medium}, and \emph{large}. The small tier contains $1{,}350$ instances, the medium tier contains $1{,}650$ instances, and the large tier contains $368$ instances. The small tier focuses primarily on $100$-task instances with moderate precedence density across empty, random, and warehouse maps. The medium tier expands to larger $200$-task instances with broader sweeps in both agent count and precedence density, together with several larger warehouse-only settings. The large tier is warehouse-only and contains the hardest instances in the benchmark suite, reaching up to $500$ agents and $1{,}000$ tasks (full configurations in Supplement~\ref{app:benchmark-configs}). 

Within each map family and tier, the benchmark suite contains multiple independently generated instances for each realized $(k,m,|P|)$ configuration. All comparisons use the common subset of instances for which every displayed method has a corresponding seeded result. All compared methods use the same fixed-assignment seed for a given instance and seed solver, so differences between methods are not confounded by different initialization quality.

\subsection{Protocol and Configuration}
\label{sec:exp-setup}

The primary comparison set consists of four methods, namely \textit{Regret}, \textit{Local--PBS}, \textit{Global--PBS}, and \textit{Global--CBS}. \textit{Regret} uses regret-based repair, while the remaining three use neighborhood MAPF-PC repair, with \textit{Local}/\textit{Global} denoting reassignment scope and \textit{PBS}/\textit{CBS} identifying the solver type. The solver type refers to the MAPF-PC solver used during repair, not the solver used to generate the initial seed. These four methods are chosen to isolate two key design axes. \textit{Regret} against \textit{Local--PBS} tests whether MAPF-PC neighborhood repair improves over greedy reinsertion, \textit{Local--PBS} against \textit{Global--PBS} tests the effect of reassignment scope, and \textit{Global--PBS} against \textit{Global--CBS} compares the underlying MAPF-PC solver under global reassignment. All four use \emph{hard} repair only, retaining only conflict-free candidates, with relaxed variants in Supplement~\ref{app:relaxed}.

Across the reported LNS runs, the shared configuration uses destroy seed size $2$, adaptive destroy selection through ALNS~\cite{alns}, and threshold acceptance~\cite{thresholdacc}. ALNS uses a reaction factor of $0.35$ with standard reward-based weight updates. Threshold acceptance uses a cooling coefficient of $0.99975$ with initial temperature set to $5\%$ of the initial sum of costs. We also evaluated alternative acceptance rules, including simulated annealing~\cite{van1987simulated}, great deluge~\cite{gda}, and old bachelor acceptance~\cite{oba}, but threshold acceptance consistently performed best across our benchmark suite. MLA* is used for local patching during neighborhood destruction (Section~\ref{sec:neighborhood-destruction}), while SIPPS serves as the low-level planner inside MAPF-PC neighborhood repair. Large-tier results are PBS-seeded only, since CBS-PC seeds were not available at that scale. Our primary objective is sum of costs (SoC). We report absolute SoC reduction $\Delta J = J_{\mathrm{seed}} - J_{\mathrm{final}}$, relative SoC reduction $100 \cdot \Delta J / J_{\mathrm{seed}}$, and improvement frequency (fraction of instances with $\Delta J > 0$). To address \textbf{Q4}, we additionally track the reduction in total precedence wait.

We distinguish explicitly between seed generation, the main LNS loop, and post-refinement. The total reported runtime includes all three components. All compared LNS runs use tier-specific wall-clock budgets of $60\,\mathrm{s}$ for the small tier, $180\,\mathrm{s}$ for the medium tier, and $300\,\mathrm{s}$ for the large tier. The wall-clock budget governs only the main LNS loop. Seed generation and post-refinement run outside this budget, and their costs are reported separately. The framework is implemented in C++ and tested on an Intel Core i9-14900K running Ubuntu 24.04.\footnote{The code and the problem instances will be released.}

\begin{figure*}[t]
    \centering
    \includegraphics[width=\textwidth]{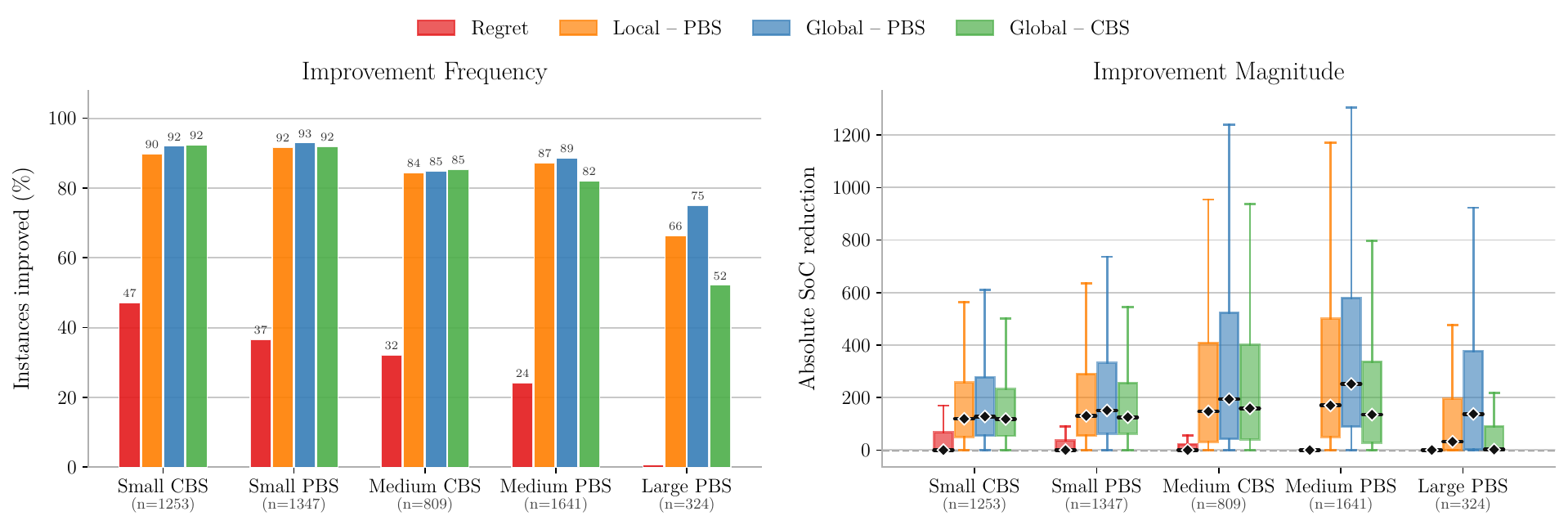}
    \caption{
        Improvement frequency and absolute sum of costs reduction. \textbf{Left}: Fraction of improved instances, \textbf{Right}: Boxplots of absolute SoC reduction across all instances. Larger values are better.
    }
    \label{fig:gain-frequency}
\end{figure*}

\section{Results}
\label{sec:results}

\subsection{Comparison of Repair Strategies}

To answer \textbf{Q1}, we compare \textit{Regret}, \textit{Local--PBS}, \textit{Global--PBS}, and \textit{Global--CBS}. Figure~\ref{fig:main-method-comparison} reports relative sum of costs reduction over the fixed-assignment seed. Across both PBS- and CBS-seeded settings, all neighborhood-repair methods substantially outperform the regret baseline. For example, \textit{Global--PBS} achieves a median relative reduction of 12.2\% overall, whereas \textit{Regret} achieves a median reduction of 0.0\%, indicating that it fails to improve the majority of instances. Among the MAPF-PC repair variants, \textit{Global--PBS} is the strongest overall. It achieves the largest median relative reduction in most small and medium settings and remains the only method with nontrivial gains on the large PBS-seeded warehouse tier, where it attains a median relative reduction of 1.5\% and improves 75.0\% of instances. \textit{Local--PBS} is consistently competitive and often close behind, while \textit{Global--CBS} is competitive in some easier settings but achieves smaller median reductions than \textit{Global--PBS} on the harder medium-tier settings.

Figure~\ref{fig:gain-frequency} shows that this advantage is not driven by a few isolated wins. In the left panel, \textit{Global--PBS} improves a large fraction of instances and leads in most benchmark subsets. In the right panel, the distribution of absolute SoC reduction is generally higher for \textit{Global--PBS}, and its median reduction is the largest in each subset shown. This is especially clear on the large PBS-seeded warehouse tier, where \textit{Global--PBS} improves 75.0\% of instances while the regret baseline remains nearly flat. These results indicate that \textit{Global--PBS} generally finds better solutions more often and by larger margins than the other configurations.

\begin{figure}[t]
    \centering
    \includegraphics[width=\columnwidth]{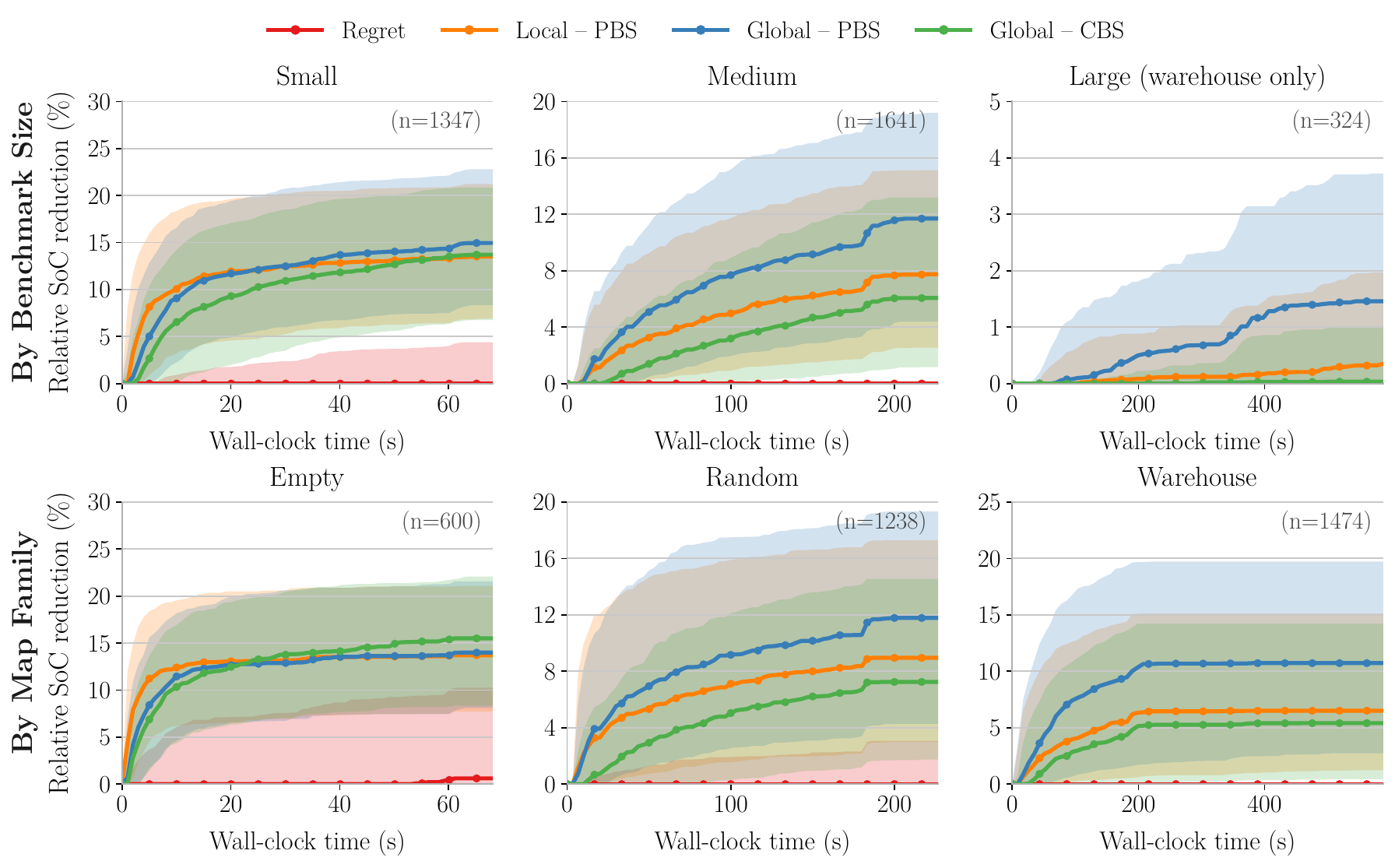}
    \caption{
    Search behavior on PBS-seeded runs. Curves show median best-so-far relative SoC reduction over wall-clock time, and shaded bands show the interquartile range. The x-axis shows wall-clock time including seed generation and post-refinement.
    }
    \label{fig:search-behavior-pbs}
\end{figure}

\subsection{Search Behavior Over Time}

To address \textbf{Q2}, Figure~\ref{fig:search-behavior-pbs} provides a complementary temporal view of these trends. The regret baseline remains close to zero over time, whereas the MAPF-PC repair methods improve rapidly once the search begins. For \textit{Global--PBS}, on small instances, the median curve reaches 50\% of its final gain by about $8\,\mathrm{s}$ and 80\% by about $24\,\mathrm{s}$. On the medium and pooled warehouse panel, it reaches 50\% by about $67\text{--}69\,\mathrm{s}$ and 80\% by about $138\text{--}160\,\mathrm{s}$. This shows that neighborhood repair quickly identifies productive reassignment opportunities. At the same time, the stronger methods continue to improve rather than plateauing immediately, which indicates that the gains are not confined to a single early repair.

A useful nuance emerges when comparing \textit{Local--PBS} and \textit{Global--PBS}. In some easier settings, especially on smaller instances, \textit{Local--PBS} improves very quickly at the start of the run. However, given more runtime and in harder settings, \textit{Global--PBS} typically catches up and then overtakes it. This is most visible in the PBS-seeded medium and warehouse subsets, where the best-so-far curve of \textit{Global--PBS} ultimately dominates. Localized repair can be attractive for rapid early progress, but globally flexible reassignment yields the strongest overall search behavior given sufficient runtime. CBS-seeded search behavior and iteration-level curves are provided in Supplement~\ref{app:search-behavior}.

\subsection{Scalability and High-Precedence Regimes}

\begin{figure}[t]
    \centering
    \includegraphics[width=\columnwidth]{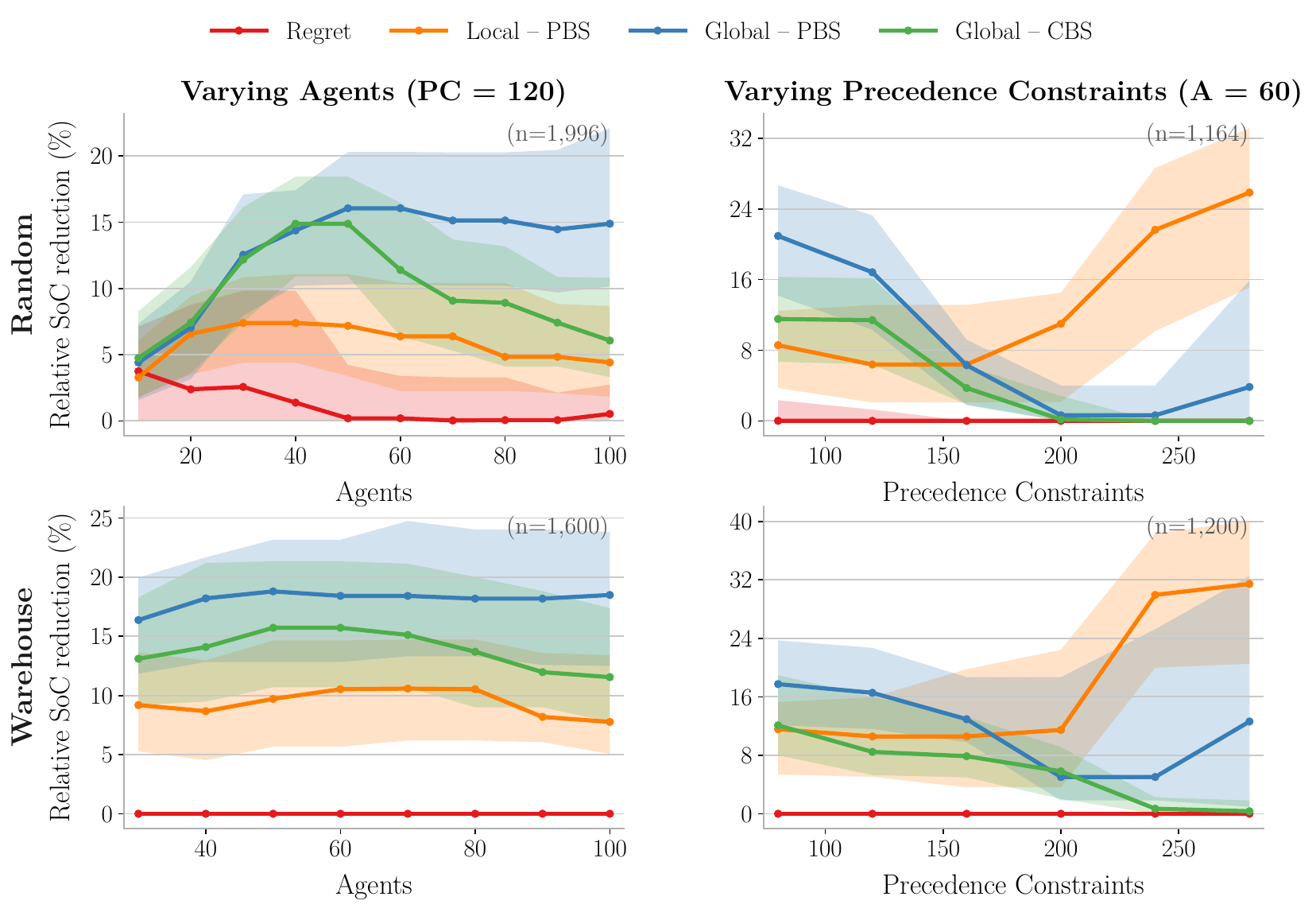}
    \caption{
    Scalability on PBS-seeded runs. Curves show median relative SoC reduction, and shaded bands show interquartile range. \textbf{Left column}: Varying agents with fixed tasks and precedence constraints. \textbf{Right column}: Varying precedence constraints with fixed agents and tasks. \textbf{Top row}: Random map. \textbf{Bottom row}: Warehouse map.
    }
    \label{fig:scalability-pbs}
\end{figure}

Turning to \textbf{Q3}, the scalability results further clarify when global flexibility is most valuable. Figure~\ref{fig:scalability-pbs} shows that when the number of agents increases while the numbers of tasks ($m = 200$) and precedence constraints ($|P| = 120$) are held fixed, \textit{Global--PBS} remains the strongest method across a broad range of settings. This indicates that broader reassignment flexibility remains beneficial as team size grows. The same pattern appears across random and warehouse environments, which suggests that the advantage of global repair is not tied to a single map family.

At low to moderate numbers of precedence constraints with $k = 60$ agents and $m = 200$ tasks, \textit{Global--PBS} continues to perform strongly. However, in the high-precedence regime, its advantage erodes and \textit{Local--PBS} performs better. Supplement~\ref{app:high-prec} provides further analysis of this crossover. As precedence density increases, the number of accepted repairs per run drops sharply for \textit{Global--PBS}, while \textit{Local--PBS} retains a more stable repair rate. This suggests that dense precedence structure makes globally flexible repair harder to realize productively, as larger neighborhoods encounter more cross-boundary precedence constraints and are more likely to produce failed or uncompetitive repairs. The more constrained local mode avoids this by keeping neighborhoods small enough to remain tractable.

\subsection{Precedence-Induced Wait Reduction}

\begin{figure}[t]
    \centering
    \includegraphics[width=\columnwidth]{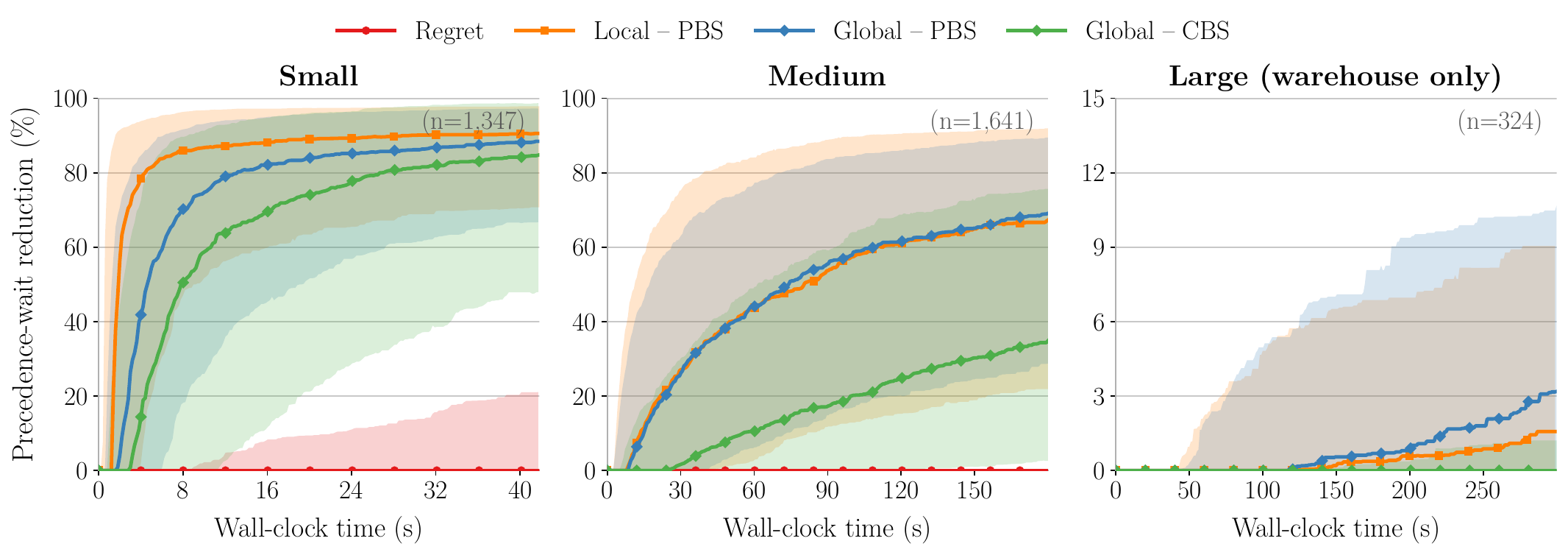}
    \caption{
    Reduction in realized precedence wait over time on PBS-seeded runs. Larger values indicate greater removal of precedence-induced waiting relative to the seed solution.
    }
    \label{fig:precedence-pbs}
\end{figure}

Finally, \textbf{Q4} asks whether the sum of costs improvements are primarily driven by reductions in precedence-induced waiting. If so, the methods that reduce precedence wait the most should also achieve the largest SoC gains. Figure~\ref{fig:precedence-pbs} tracks the reduction in realized precedence wait over time. On PBS-seeded small instances, \textit{Local--PBS} reduces precedence wait especially quickly, with \textit{Global--PBS} close behind. On medium instances, the two PBS-based methods remain close, while \textit{Global--CBS} is weaker. On the large PBS-seeded subset, only \textit{Global--PBS} produces a noticeable reduction. These curves confirm that the precedence-aware aspects of the neighborhood design are doing meaningful work. However, on small instances \textit{Local--PBS} reduces precedence wait faster yet does not achieve a better final SoC than \textit{Global--PBS}, and on the large tier \textit{Global--PBS} leads in both metrics. This suggests that relieving precedence bottlenecks is an important but incomplete explanation of the overall gains, since globally flexible reassignment can also improve routing efficiency by placing tasks with agents that reach them at lower path cost.

Additional analyses, including operator-contributions, relaxed soft-conflict variants, and a runtime decomposition by tier and method, are provided in the Supplement.

\section{Conclusion}
We studied TAPF-PC, a lifted setting in which assignment, ordering, and collision-free routing must be decided jointly under precedence constraints. We argued that fixed-assignment MAPF-PC does not resolve this larger problem, because allowing reassignment enlarges the solution space and creates improvement opportunities that are inaccessible once task ownership is fixed. To address this, we developed a LNS framework that separates the problem into an outer reassignment search and inner MAPF-PC repair subproblems, with SIPPS integrated as an efficient low-level planner.

Our experiments show that this approach improves 89.1\% of instances over fixed-assignment seeds, with the threshold-accepting \textit{Global--PBS} variant performing best overall. More broadly, the results suggest that MAPF-PC methods are especially effective for TAPF-PC when used as repair engines inside a higher-level search over flexible assignment. At the same time, the approach is heuristic, and the high-precedence regime reveals a limitation of globally flexible repair, as \textit{Local--PBS} can become more effective when dense precedence structure makes global repair harder to exploit.

Despite these encouraging results, several directions remain for future work. One is to design stronger repair policies for denser-precedence regimes. Another is to extend the framework to lifelong settings where tasks arrive online and precedence relations evolve over time.

\appendix

\section*{Acknowledgments}
We gratefully acknowledge support from the Boeing Company. We thank Siddhant Tandon for  help with the experimental setup. The views and conclusions contained in this document are those of the authors and should not be interpreted as representing the official policies, either expressed or implied, of the sponsoring organizations or agencies.

\section{Benchmark Configurations}
\label{app:benchmark-configs}

Table~\ref{tab:benchmark-configs} lists the full set of realized benchmark configurations by tier. Each configuration contains 50 independently generated instances.

\begin{table*}[t]
\centering
\caption{Benchmark configurations by tier. Each tuple $(k,m,|P|)$ denotes the number of agents, tasks, and precedence constraints. Each configuration contains 50 independently generated instances.}
\label{tab:benchmark-configs}
\small
\setlength{\tabcolsep}{6pt}
\renewcommand{\arraystretch}{1.3}
\begin{tabular}{p{1.2cm} p{2.8cm} c p{8.8cm}}
\toprule
\textbf{Tier} & \textbf{Map families} & \textbf{\#} & \textbf{Realized} $(k,m,|P|)$ \textbf{configurations} \\
\midrule
\rowcolor{lightgray}
Small
&
\texttt{empty-16}\newline
\texttt{empty-32}\newline
\texttt{random-32}\newline
\texttt{warehouse}$^\dagger$
& 8 &
$(10,100,80)$, $(10,100,100)$, $(20,100,80)$, $(20,100,100)$, $(30,100,80)$, $(30,100,100)$, $(60,200,80)$, $(100,200,100)$
\\[2pt]
Medium
&
\texttt{random-32}\newline
\texttt{warehouse}$^\dagger$
& 21 &
$(10,200,120)$, $(10,200,160)$, $(10,200,280)$, $(20,200,120)$, $(20,200,160)$, $(20,200,280)$, $(30,200,120)$, $(40,200,120)$, $(50,200,120)$, $(60,200,120)$, $(60,200,160)$, $(60,200,200)$, $(60,200,240)$, $(60,200,280)$, $(70,200,120)$, $(80,200,120)$, $(90,200,120)$, $(100,200,120)$, $(150,300,150)$, $(200,400,200)$, $(250,500,250)$
\\[2pt]
\rowcolor{lightgray}
Large
&
\texttt{warehouse}$^\dagger$
& 10 &
$(200,800,400)$, $(200,1{,}000,500)$, $(200,1{,}200,600)$, $(200,1{,}400,700)$, $(200,1{,}600,800)$, $(300,600,300)$, $(350,700,350)$, $(400,800,400)$, $(450,900,450)$, $(500,1{,}000,500)$
\\
\bottomrule
\end{tabular}

\vspace{4pt}
\hfill{\footnotesize $^\dagger$\texttt{warehouse-10-20-10-2-1}}
\end{table*}

\section{Additional Search Behavior}
\label{app:search-behavior}

The main text reports search behavior over wall-clock time for PBS-seeded runs. This section provides complementary views.

\subsection{CBS-Seeded Search Behavior}

\begin{figure}[t]
    \centering
    \includegraphics[width=\columnwidth]{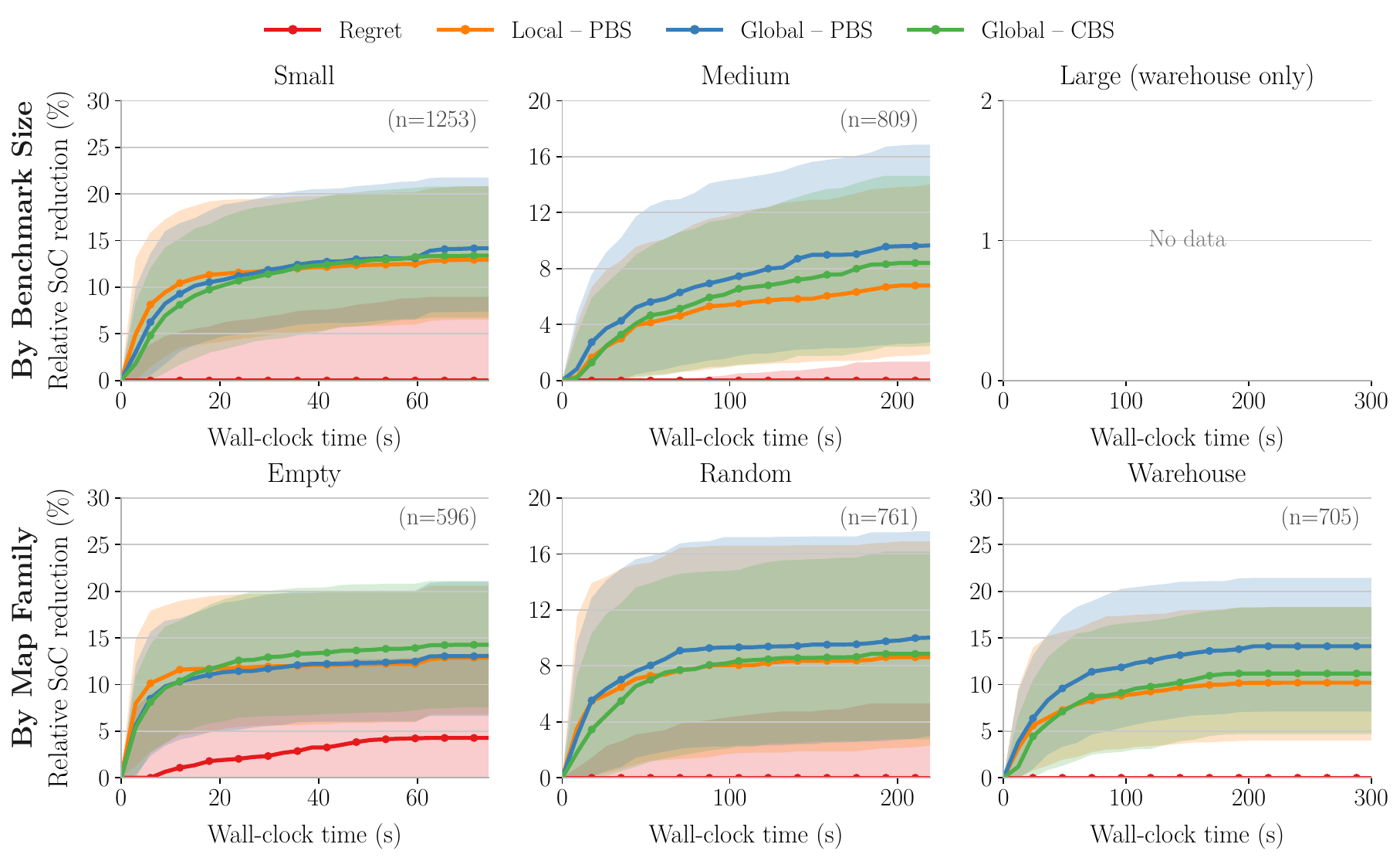}
    \caption{
    Search behavior on CBS-seeded runs. Curves show median best-so-far relative SoC reduction over wall-clock time, and shaded bands show the interquartile range. No CBS-seeded data is available for the large tier.
    }
    \label{fig:search-behavior-cbs}
\end{figure}

Figure~\ref{fig:search-behavior-cbs} shows search behavior over wall-clock time for CBS-seeded runs. The overall trends are broadly consistent with the PBS-seeded results. On small and medium instances, the MAPF-PC repair methods outperform the regret baseline, and \textit{Global--PBS} again emerges as the strongest configuration. One notable difference is that \textit{Regret} achieves non-negligible gains on CBS-seeded empty maps, suggesting that higher-quality seeds provide a more favorable starting point for greedy reinsertion. No CBS-seeded data is available for the large tier, as CBS-PC did not scale to that tier.

\subsection{Iteration-Level Search Behavior}

\begin{figure*}[t]
    \centering
    \includegraphics[width=\textwidth]{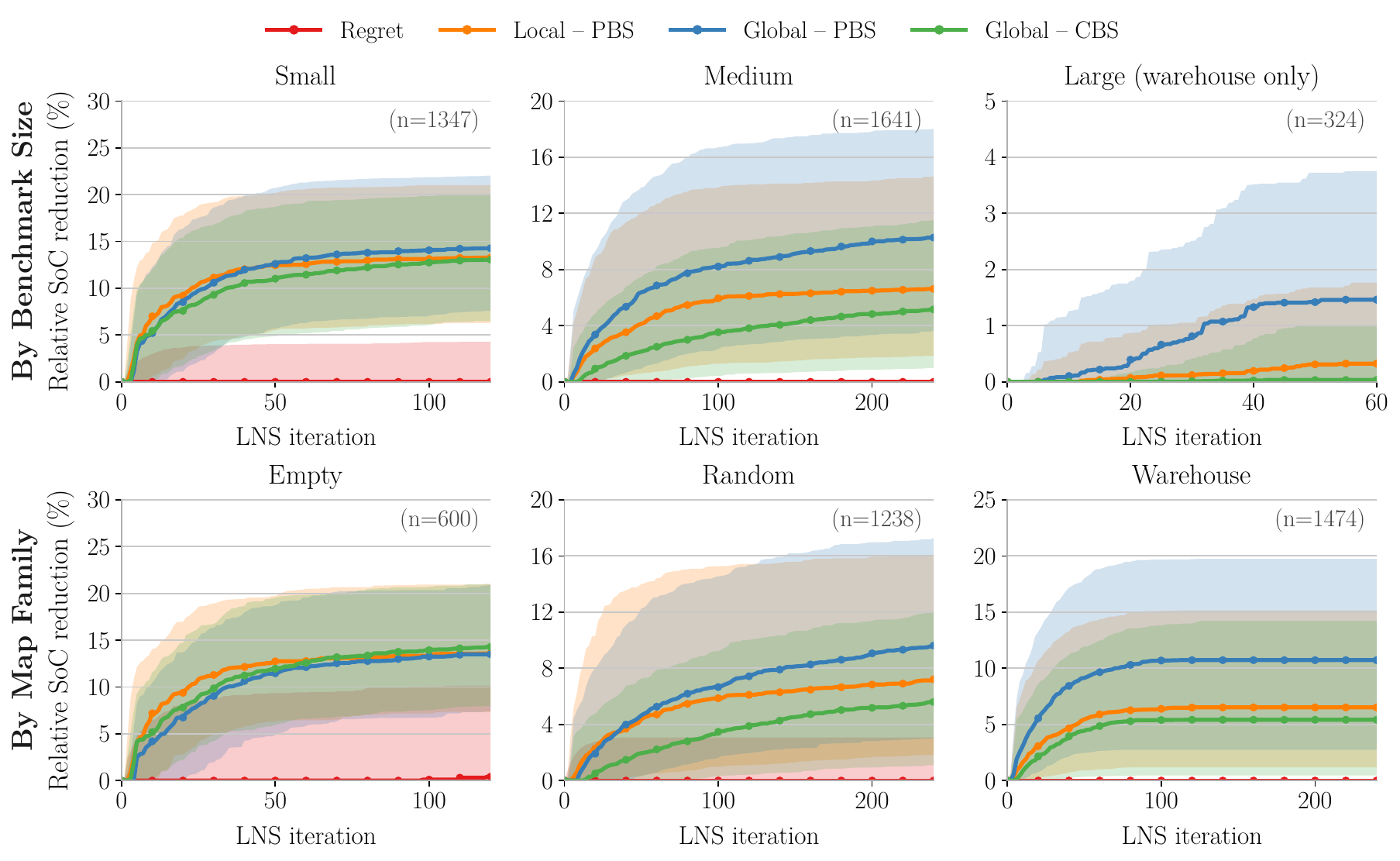}
    \caption{
    Search behavior by LNS iteration on PBS-seeded runs. Curves show median best-so-far relative SoC reduction over iteration count, and shaded bands show the interquartile range.
    }
    \label{fig:search-behavior-iter-pbs}
\end{figure*}

\begin{figure*}[t]
    \centering
    \includegraphics[width=\textwidth]{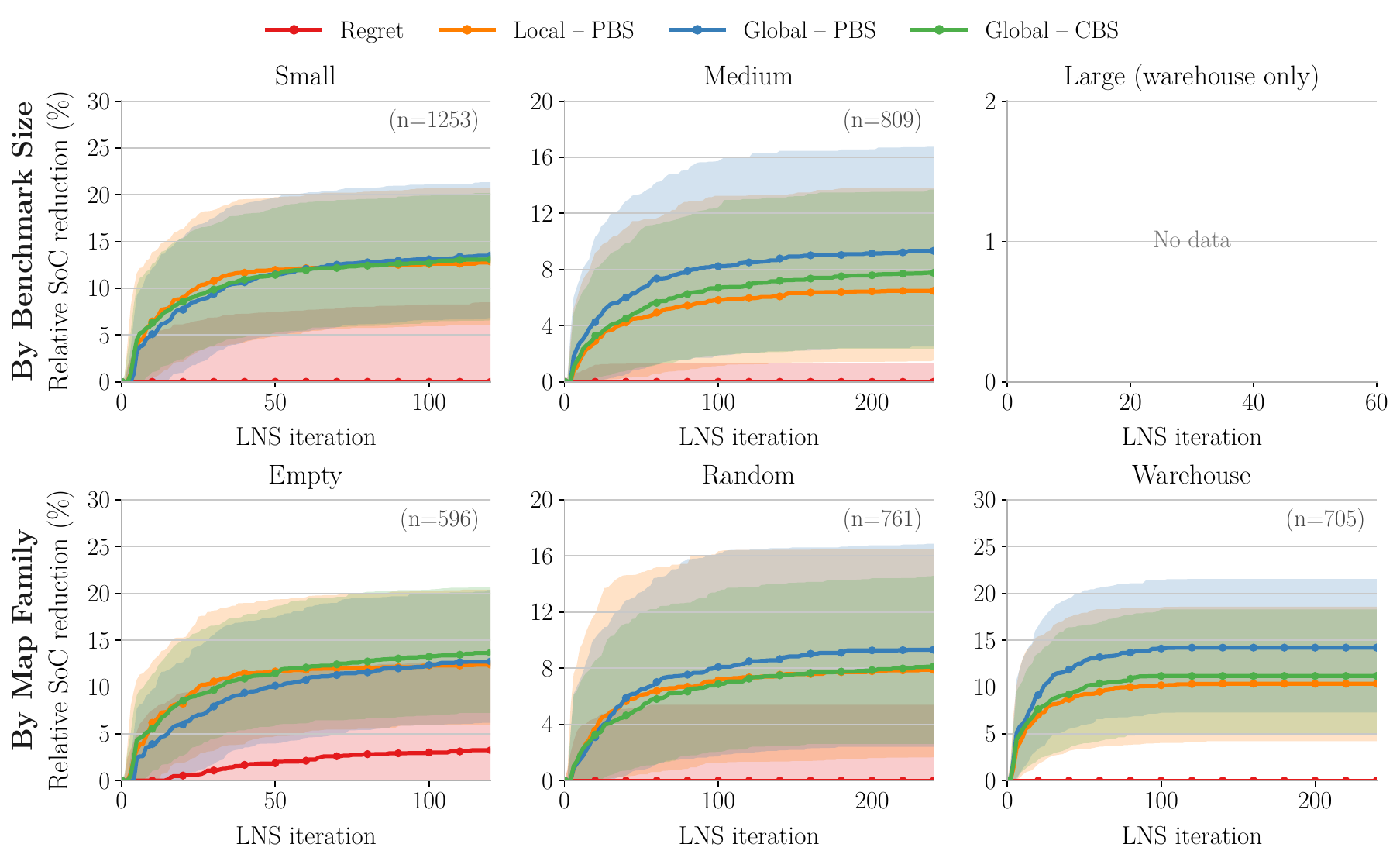}
    \caption{
    Search behavior by LNS iteration on CBS-seeded runs. Curves show median best-so-far relative SoC reduction over iteration count, and shaded bands show the interquartile range. No CBS-seeded data is available for the large tier.
    }
    \label{fig:search-behavior-iter-cbs}
\end{figure*}

Figures~\ref{fig:search-behavior-iter-pbs} and~\ref{fig:search-behavior-iter-cbs} show search behavior as a function of LNS iteration count rather than wall-clock time, for PBS-seeded and CBS-seeded runs respectively. This view shows how much each method improves per iteration, independent of how long each iteration takes. On the PBS-seeded medium and large tiers, \textit{Global--PBS} achieves larger per-iteration gains than \textit{Local--PBS}, confirming that its wall-clock advantage stems from higher-quality repairs. The CBS-seeded iteration curves show a similar pattern on small and medium instances.

\section{Relaxed Soft-Conflict Variants}
\label{app:relaxed}

The primary results use hard repair, in which only fully conflict-free neighborhood realizations are retained. This section compares hard and relaxed variants for both local and global PBS-based repair.

\begin{figure*}[t]
    \centering
    \includegraphics[width=\textwidth]{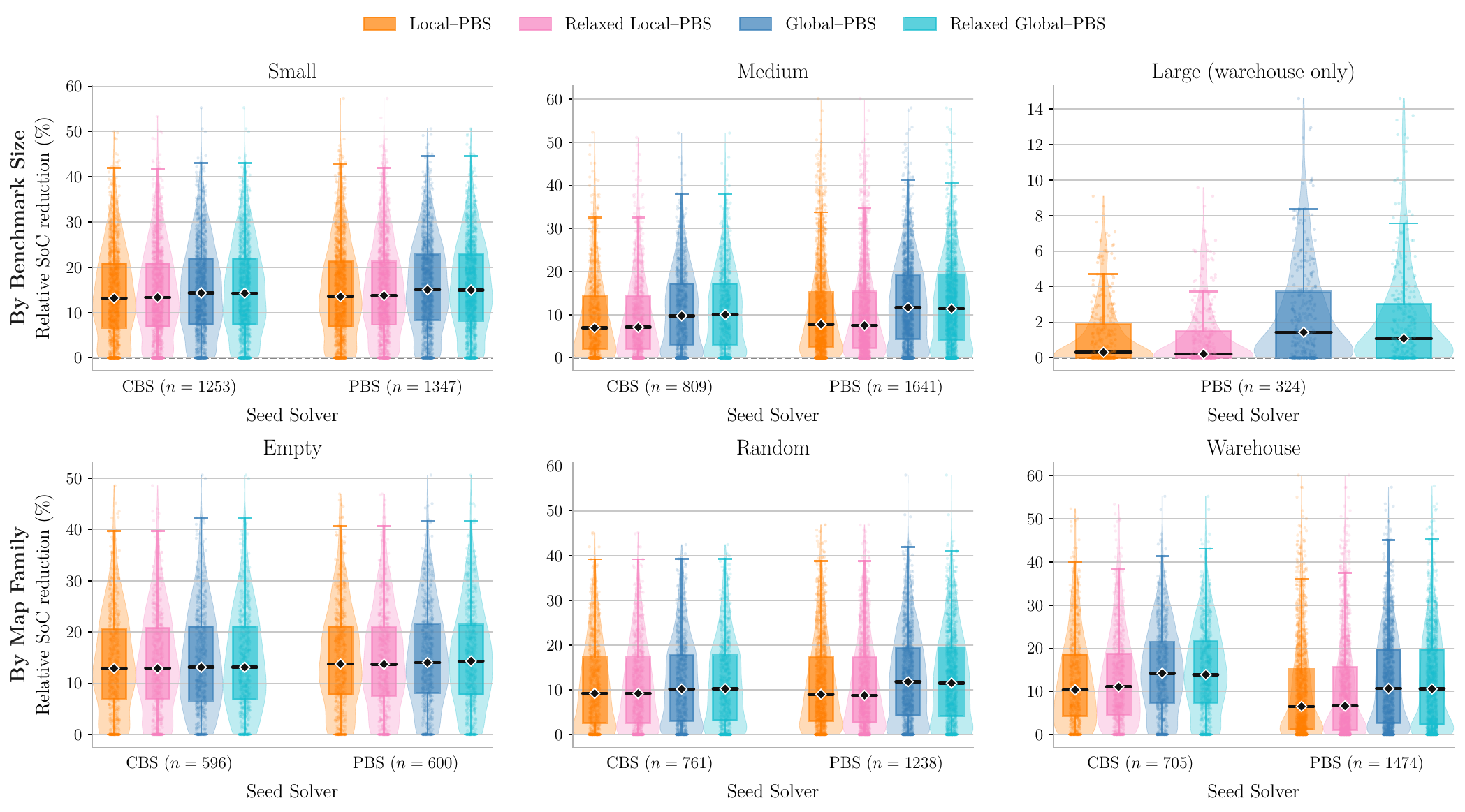}
    \caption{Hard vs.\ relaxed repair comparison across benchmark subsets. Each panel reports relative sum-of-costs reduction over the fixed-assignment seed. Diamond markers indicate the median. \textbf{Top row}: By benchmark tier. \textbf{Bottom row}: By map family.}
    \label{fig:relaxed-comparison}
\end{figure*}

\begin{figure*}[t]
    \centering
    \includegraphics[width=\textwidth]{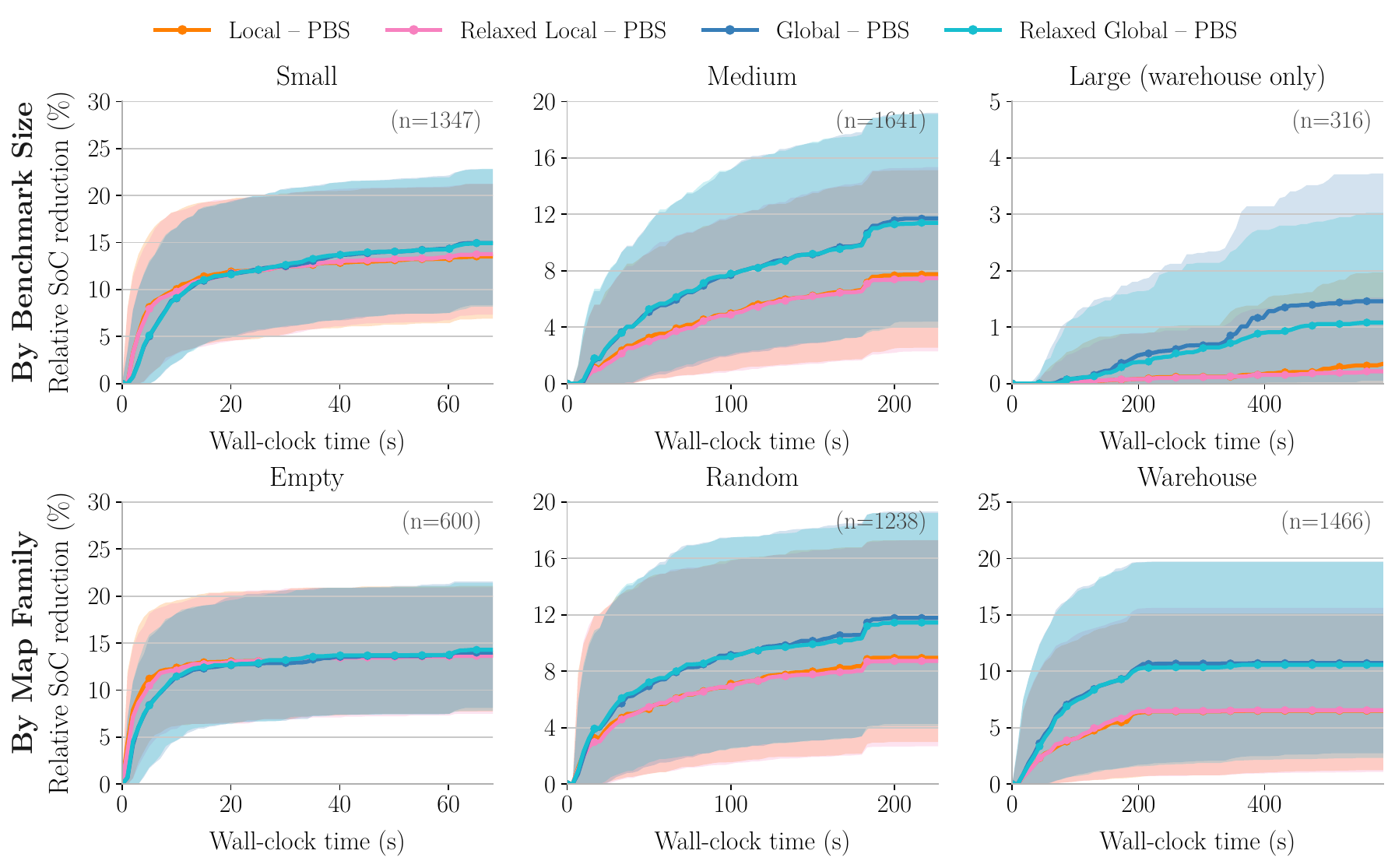}
    \caption{Search behavior for hard vs.\ relaxed PBS variants. Curves show median best-so-far relative SoC reduction over wall-clock time, and shaded bands show the interquartile range.}
    \label{fig:relaxed-search-behavior}
\end{figure*}

Figure~\ref{fig:relaxed-comparison} shows that the effect of relaxed repair is limited. On small and medium instances, the hard and relaxed variants perform similarly for both local and global modes. On the large tier, hard \textit{Global--PBS} retains a higher median reduction than its relaxed counterpart, indicating that the overhead of resolving accumulated soft conflicts can outweigh the benefit of accepting more candidates on the hardest instances. For \textit{Local--PBS}, the hard and relaxed variants perform similarly throughout, with neither consistently dominating. Figure~\ref{fig:relaxed-search-behavior} provides a temporal view of the same comparison. The local variants track each other closely across all tiers. The global variants also remain close on small and medium instances, but on the large tier hard \textit{Global--PBS} maintains a clear lead throughout.

These results justify the choice of hard \textit{Global--PBS} as the primary configuration. While relaxed repair performs comparably on small and medium instances, it does not improve performance on the hardest instances.

\section{High-Precedence Diagnostic}
\label{app:high-prec}

\begin{figure}[t]
    \centering
    \includegraphics[width=\columnwidth]{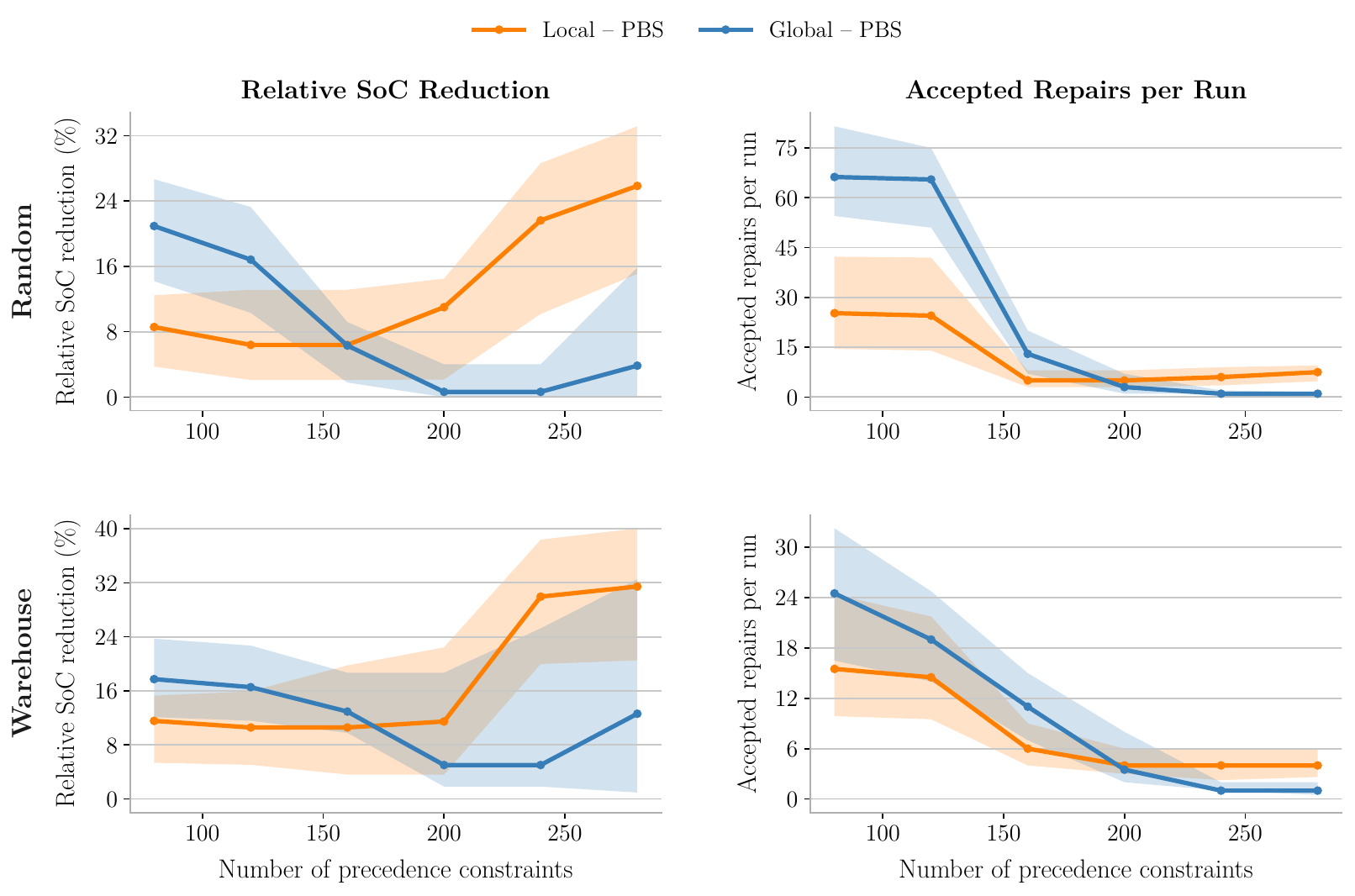}
    \caption{
    PBS-seeded high-precedence diagnostic. The left column shows median relative sum-of-costs reduction, and the right column shows median accepted repairs per run. Shaded bands show the interquartile range. The top row corresponds to random maps, and the bottom row corresponds to warehouse maps. All instances use $k=60$ agents and $m=200$ tasks.
    }
    \label{fig:scalability-highpc}
\end{figure}

Figure~\ref{fig:scalability-highpc} provides a detailed view of the crossover between \textit{Global--PBS} and \textit{Local--PBS} in the high-precedence regime, as discussed in Section~6.3 of the main text. As precedence density increases, the number of accepted repairs per run drops sharply for \textit{Global--PBS}, while \textit{Local--PBS} retains a more stable level of accepted repairs. This indicates that globally flexible repair becomes harder to realize productively when precedence structure is extremely dense, as larger destroyed neighborhoods encounter more cross-boundary precedence constraints and are more likely to produce failed or uncompetitive repair candidates. The more constrained local mode avoids this by keeping neighborhoods small enough to remain tractable. Overall, \textit{Global--PBS} remains the strongest method across most regimes. However, very high precedence density presents a genuine challenge where local repair can become the better choice.

\section{Runtime Decomposition}
\label{app:runtime}

\begin{figure}[t]
    \centering
    \includegraphics[width=\columnwidth]{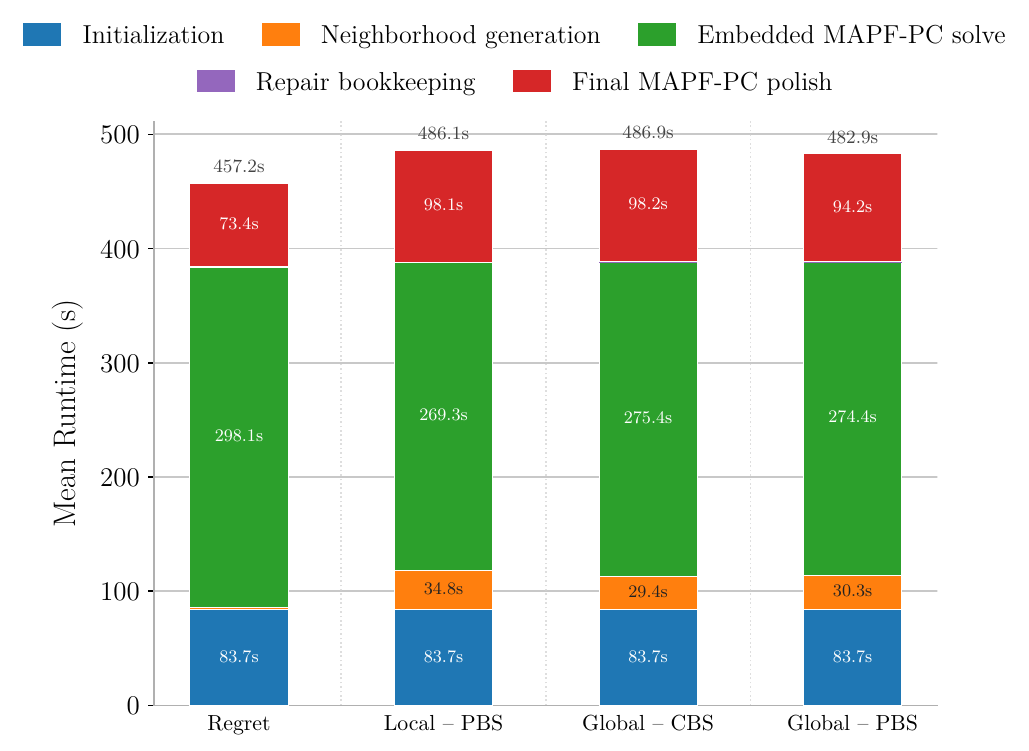}
    \caption{
    Runtime decomposition on the large PBS-seeded warehouse tier. The embedded MAPF-PC solve dominates the total runtime, while neighborhood generation, bookkeeping, and final polishing remain secondary.
    }
    \label{fig:runtime-breakdown-large}
\end{figure}

Figure~\ref{fig:runtime-breakdown-large} shows the runtime decomposition on the large PBS-seeded warehouse tier. The majority of the wall-clock budget is spent inside the embedded MAPF-PC repair solve, with the reassignment-and-order proposal stage, initialization, and neighborhood generation accounting for smaller fractions. This allocation reflects the design of the framework, as the quality gains come from solving full MAPF-PC subproblems rather than from greedy insertion, so dedicating most of the budget to the repair solve is where the investment pays off. Final MAPF-PC polish is non-negligible but still secondary. Notably, the search-behavior curves in the main text show substantial improvement well before the end of the allotted runtime, confirming that the central gains arise from the neighborhood-search process itself and that post-refinement serves as a polishing step rather than the main source of improvement. The same general pattern holds on the medium tier, though the relative share of the embedded solve is smaller. On the small tier, the breakdown is more balanced across components.

\section{Operator Contributions}
\label{app:operator-contributions}

\begin{figure*}[t]
    \centering
    \includegraphics[width=\textwidth]{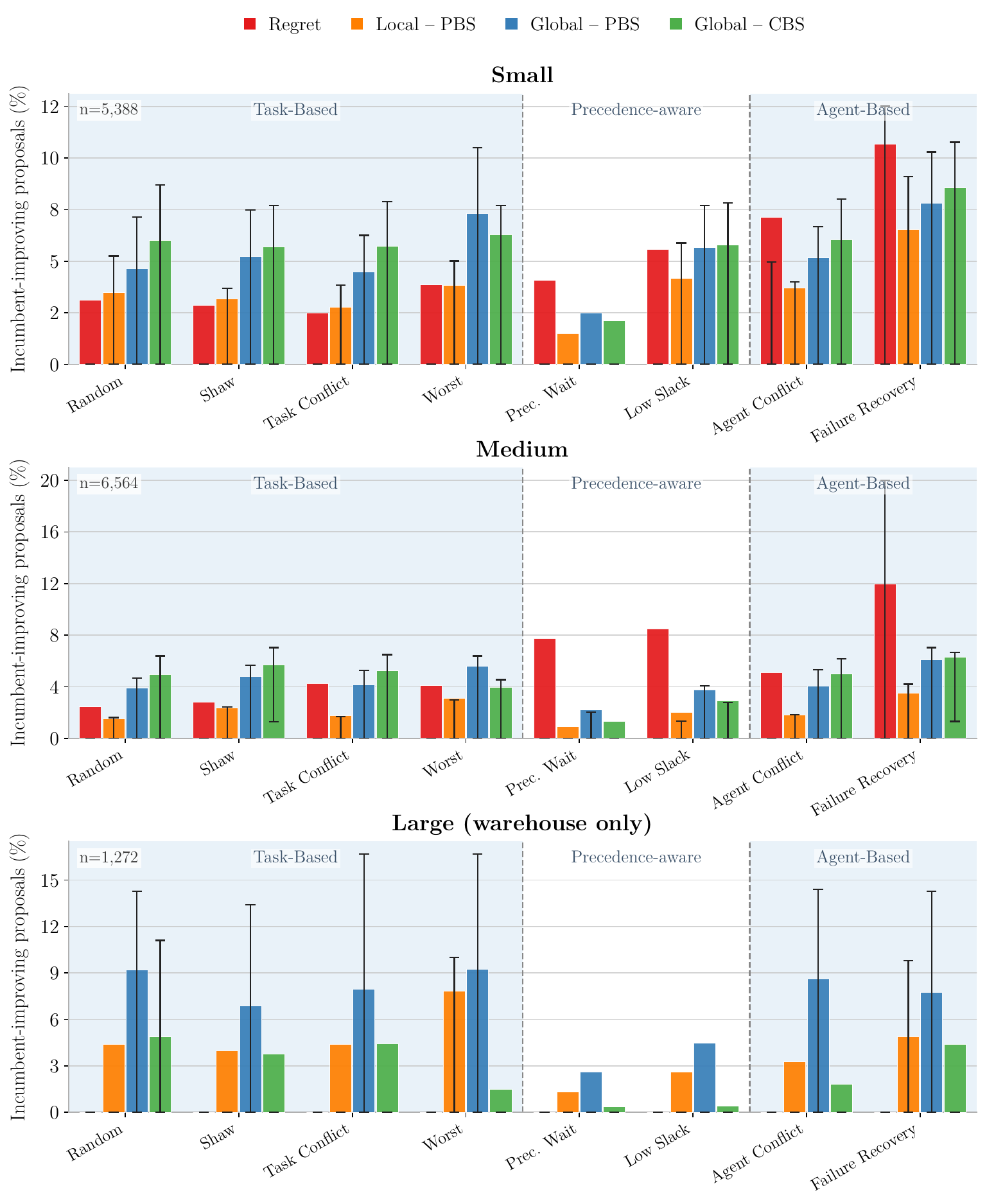}
    \caption{Destroy-operator contribution by benchmark tier on PBS-seeded runs. For each destroy operator, the colored bars show the mean incumbent-improving proposal rate (\%) aggregated over the runs in that tier, and the vertical whiskers indicate the 25th--75th percentile range of the corresponding per-run rates. Operators are grouped by family. Note that $n$ denotes the total number of method-runs contributing to the tier panel, summed across methods.}
    \label{fig:operator-contributions-pbs}
\end{figure*}

\begin{figure*}[t]
    \centering
    \includegraphics[width=\textwidth]{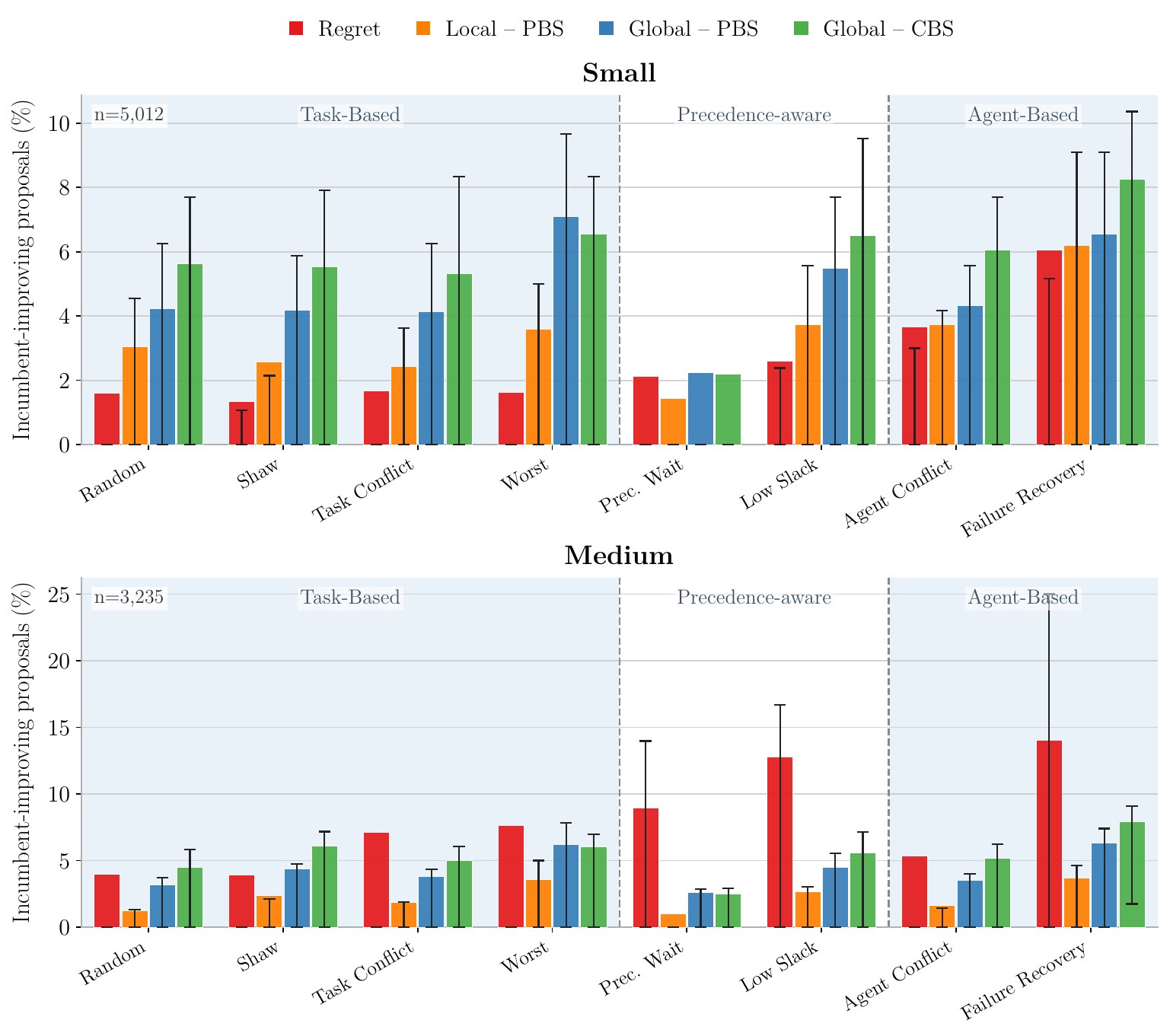}
    \caption{Destroy-operator contribution by benchmark tier on CBS-seeded runs. Format follows Figure~\ref{fig:operator-contributions-pbs}. No CBS-seeded data is available for the large tier.}
    \label{fig:operator-contributions-cbs}
\end{figure*}

Figures~\ref{fig:operator-contributions-pbs} and~\ref{fig:operator-contributions-cbs} report the incumbent-improvement rate of each destroy operator across the four methods, broken down by benchmark tier for PBS-seeded and CBS-seeded runs respectively.

The \textsc{Failure-Recovery} operator consistently achieves the highest improvement rate across all methods and tiers, reflecting the value of redirecting the search after unsuccessful repair attempts. Among the task-based operators, Worst and Shaw tend to produce slightly higher improvement rates than Random and Task Conflict, though all four contribute throughout. The precedence-aware operators show a split. \textsc{Low-Slack} achieves moderate improvement rates comparable to the task-based family, while \textsc{Precedence-Wait} has among the lowest rates across all non-Regret methods. This does not mean \textsc{Precedence-Wait} is unproductive, as it may steer the search toward neighborhoods that enable later improvements through other operators, but its direct contribution to incumbent updates is smaller.

These results support the use of a mixed ALNS portfolio. No single operator family dominates, and the combination of task-based, precedence-aware, and agent-based operators provides complementary signals for neighborhood selection.

\bibliography{aaai2026}

\end{document}